\documentclass[10pt,journal,compsoc]{IEEEtran}
\ifCLASSOPTIONcompsoc
  \usepackage[nocompress]{cite}
\else
  \usepackage{cite}
\fi
\ifCLASSINFOpdf
\else
\fi
\usepackage{url}

\usepackage{algorithm}
\usepackage{algorithmic}
\usepackage{multirow}
\usepackage{latexsym}
\usepackage{subfigure}
\usepackage{graphicx}
\usepackage{amsmath}
\usepackage{amssymb}
\usepackage{hyperref}
\hypersetup{hypertex=true,
	colorlinks=true,
	linkcolor=red,
	anchorcolor=red,
	citecolor=green}
\usepackage{color}

\begin{document}
%
\title{Prototype Completion for Few-Shot Learning}
%
%
%
%

\author{Baoquan~Zhang,
        Xutao~Li,
        Yunming~Ye,
        and~Shanshan~Feng
\IEEEcompsocitemizethanks{\IEEEcompsocthanksitem Baoquan Zhang, Xutao Li, Yunming Ye, and Shanshan Feng are with the School of Computer Science and Technology, Harbin Institute of Technology, Shenzhen,
Shenzhen 518055, Guangdong, China.\protect\\
E-mail: zhangbaoquan@stu.hit.edu.cn, \{lixutao, yeyunming\}@hit.edu.cn, victor\_fengss@foxmail.com
\IEEEcompsocthanksitem Corresponding authors are Xutao Li and Yunming Ye.\protect\\
}
\thanks{Manuscript received August 11, 2021.}}

%
%

\markboth{Journal of \LaTeX\ Class Files,~Vol.~14, No.~8, August~2015}%
{Shell \MakeLowercase{\textit{et al.}}: Bare Demo of IEEEtran.cls for Computer Society Journals}
%



\IEEEtitleabstractindextext{%
\begin{abstract}
Few-shot learning aims to recognize novel classes with few examples. Pre-training based methods effectively tackle the problem by pre-training a feature extractor and then fine-tuning it through the nearest centroid based meta-learning. However, results show that the fine-tuning step makes marginal improvements. In this paper, 1) we figure out the reason, \emph{i.e.}, in the pre-trained feature space, the base classes already form compact clusters while novel classes spread as groups with large variances, which implies that fine-tuning feature extractor is less meaningful; 2) instead of fine-tuning feature extractor, we focus on estimating more representative prototypes. Consequently, we propose a novel prototype completion based meta-learning framework. This framework first introduces primitive knowledge (\emph{i.e.}, class-level part or attribute annotations) and extracts representative features for seen attributes as priors. Second, a part/attribute transfer network is designed to learn to infer the representative features for unseen attributes as supplementary priors. Finally, a prototype completion network is devised to learn to complete prototypes with these priors. Moreover, to avoid the prototype completion error, we further develop a Gaussian based prototype fusion strategy that fuses the mean-based and completed prototypes by exploiting the unlabeled samples. Extensive experiments show that our method: (\romannumeral1) obtains more accurate prototypes; (\romannumeral2) achieves superior performance on both inductive and transductive FSL settings. Our codes are open-sourced at \url{https://github.com/zhangbq-research/Prototype_Completion_for_FSL}.

\end{abstract}

}

\maketitle

\IEEEdisplaynontitleabstractindextext

%
\IEEEpeerreviewmaketitle

\IEEEraisesectionheading{\section{Introduction}\label{sec:introduction}}

%
%
%
%
\IEEEPARstart{H}{umans} can adapt to a novel task from only a few observations, because our brains have the excellent capability of learning to learn. In contrast, modern artificial intelligence (AI) systems generally require a large amount of annotated samples to make the adaptations, such as image classification \cite{HeZRS16}. However, preparing sufficient annotated samples is often laborious, expensive, or even unrealistic in some applications such as cold-start recommendation \cite{VartakTMBL17} and drug discovery \cite{Altae-TranRPP16}. To equip the AI systems with such human-like ability, few-shot learning (FSL) becomes an important and widely studied problem. Different from conventional machine learning, FSL aims to learn a classifier from a set of base classes with abundant labeled samples, then adapt to a set of novel classes with few examples \cite{WangYKN20}. 

Existing studies on FSL roughly fall into four categories, namely the metric-based methods \cite{HaoHCWCT19, OreshkinLL18, LiLX0019}, optimization-based methods \cite{finn2017model, YaoWHL19}, graph-based methods \cite{satorras2018few, RodriguezLDL20}, and semantics-based methods \cite{xing2019adaptive, zhangLearn2021}. Though their methodologies are quite different, almost all methods address the FSL problem by a two-phase meta-learning framework, \emph{i.e.}, (\romannumeral1) a meta-training phase that learns meta-knowledge from a large number of base class tasks, and (\romannumeral2) a meta-test phase that quickly constructs a model for novel class prediction with the meta-knowledge. Recently, Chen \emph{et al. }\cite{chen2020new} find that introducing an extra pre-training phase can significantly boost the performance. In this method, a feature extractor first is pre-trained by learning a classifier on the entire base classes. Then, the metric-based meta-learning is adopted to fine-tune it. In the meta-test phase, the mean-based prototypes are constructed to classify novel classes via the nearest neighbor classifier with cosine distance. 

\begin{figure}[!t]
	\centering
	\subfigure[Base Classes ($\mathrm{\sigma^2=0.086}$)]{ 
		\label{fig1a} 
		\includegraphics[width=0.48\columnwidth]{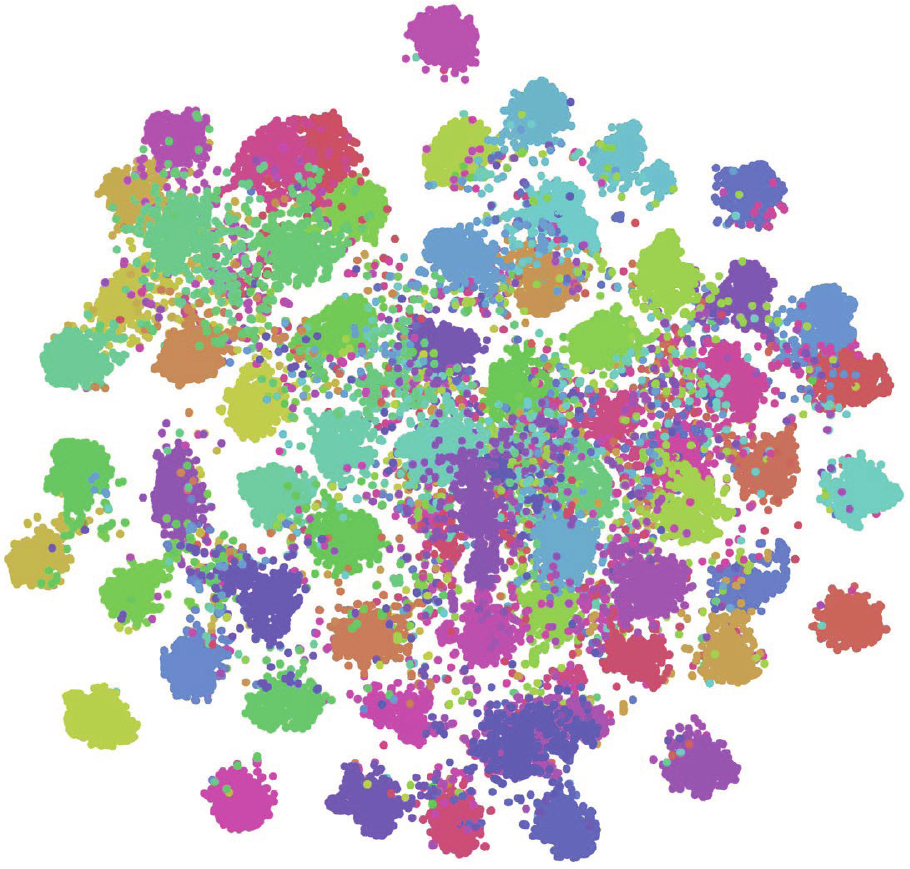}}
	\subfigure[Novel Classes ($\mathrm{\sigma^2=0.099}$)]{ 
		\label{fig1b} 
		\includegraphics[width=0.48\columnwidth]{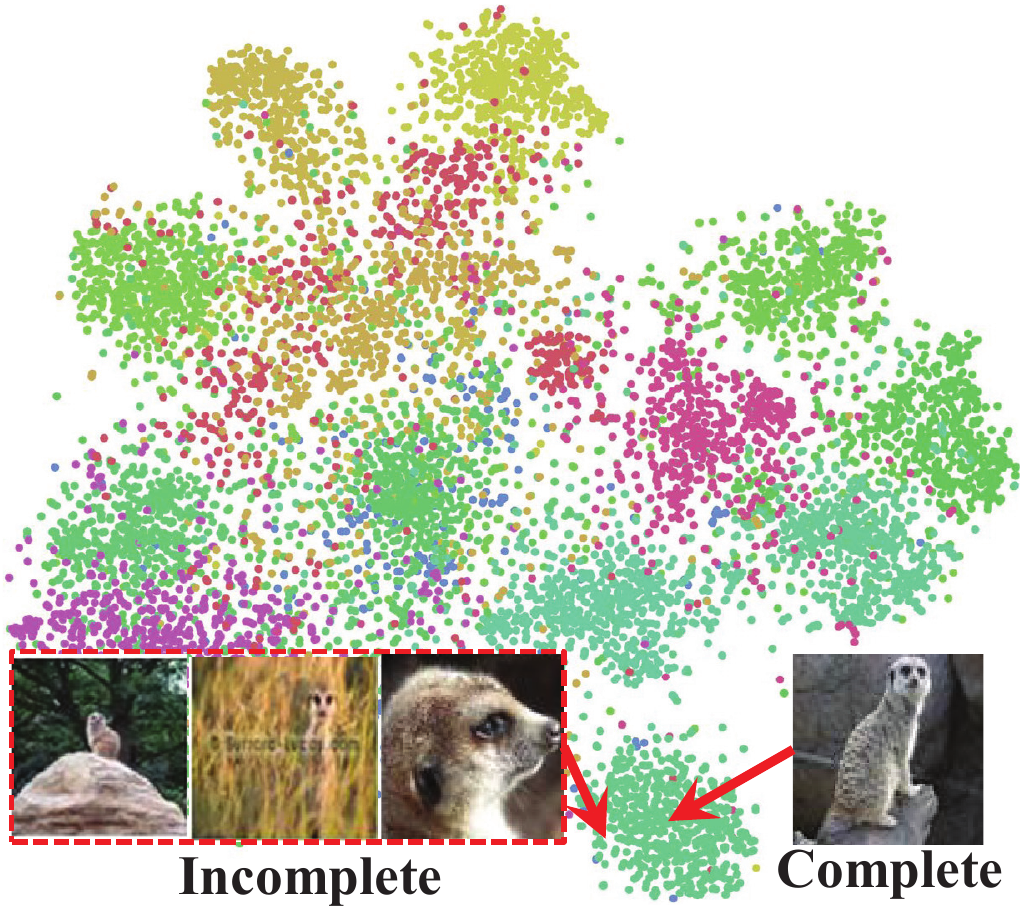}}
	\caption{The distribution of base and novel class samples of miniImagenet in the pre-trained feature space. ``$\mathrm{\sigma}^2$'' denotes the averaged variance. }
	\label{fig1}
\end{figure}

Though the pre-training based meta-learning method has achieved promising improvements on FSL, Chen \emph{et al.} find that the fine-tuning step indeed makes very marginal contributions \cite{chen2020new} during meta-learning. In other words, the power of the pre-trained model is not effectively explored by the meta-learning methods. However, the reason is not revealed in \cite{chen2020new}. To figure out the reason, we visualize the distribution of base and novel class samples of the miniImagenet in the pre-trained feature space, which is shown in Fig.~\ref{fig1}. We find that the base class samples form compact clusters while the novel class samples spread as groups with large variances. It means that (\romannumeral1) fine-tuning the feature extractor to gather the base class samples into more compact clusters is less meaningful, because this enlarges the probability to overfit the base tasks; and (\romannumeral2) the given few labeled samples may be far away from its ground-truth class centers in the case of large variances for novel classes, which poses a great challenge for estimating representative prototypes. Hence, in this paper, instead of fine-tuning feature extractor, we focus on {\bf how to estimate representative prototypes from few labeled samples}, especially when these samples are far away from their ground-truth class centers. 

Recently, Xue \emph{et al.} \cite{XueW20} also attempt to address a similar problem by learning a mapping function from noisy samples to their ground-truth class centers. However, learning to recover representative prototypes from noisy samples without any priors is very difficult. Moreover, the method does not leverage the pre-training strategy. Thus, its performance improvement is limited. In this paper, inspired by the visual attribute learning \cite{multilabelobjectattribute, WanCLYZY019}, we find that the samples deviated from its ground-truth centers are often incomplete, \emph{i.e.}, missing some representative attribute features. As shown in Fig.~\ref{fig1b}, the meerkat sample nearby the class center contains all the representative features, \emph{e.g.}, the head, body, legs, and tail, while the ones far away may miss some representative features such as legs and tail. This means that the prototypes estimated by the samples deviated from its class centers may be incomplete, which limits the classification performance of FSL. 

Based on this fact, we propose a novel prototype completion framework for FSL. Our framework works in a pre-training manner and introduces some primitive knowledge (\emph{i.e.}, class-level attribute or part annotations), e.g., whether a class object should have ears, legs or eyes, as priors to achieve the prototype completion. Specifically, we first extract the visual features for each seen part/attribute, by aggregating the pre-trained feature representations of all the base class samples that have the corresponding attributes in our primitive knowledge. Second, a \underline{P}art/\underline{A}ttribute \underline{T}ransfer \underline{Net}work (PATNet) is then designed to infer the visual features for each unseen part/attribute. Third, we mimic the setting of few-shot classification task and construct a set of prototype completion tasks. A \underline{Proto}type \underline{Com}pletion \underline{Net}work (ProtoComNet) is then developed to learn to complete representative prototypes with the primitive knowledge and the obtained visual attribute features. To avoid the prototype completion error caused by primitive knowledge noises or base-novel class differences, 
we further design a Gaussian-based prototype fusion strategy, which effectively combines the mean-based and completed prototypes by exploiting the unlabeled data. Finally, the few-shot classification is achieved via a nearest neighbor classifier. Our main contributions of this paper can be summarized as follows: 

\begin{itemize}
	\item We reveal the reason why the feature extractor fine-tuning step contributes very marginally to the pre-training based meta-learning methods, and point out that representative prototype estimation is a more important issue.
	
	\item We propose a novel prototype completion based FSL framework. In the framework, a part/attribute transfer network, a prototype completion network and a Gaussian-based prototype fusion strategy are designed, which offer our framework the excellent ability to construct more representative prototypes, by exploiting the primitive knowledge of both seen and unseen parts/attributes.
	
	\item In the Gaussian-based prototype fusion strategy, we propose and extend three methods to estimate prototype fusion parameters, i.e., a two-step estimation method, an EM (Expectation Maximization)-based estimation method, and an improved EM-based estimation method, which fully exploit the unlabeled data for more accurate prototypes estimation.
	
	\item We conduct comprehensive experiments on three real-world data sets. The experimental results demonstrate that our method achieves superior performance in both inductive and transductive FSL settings over state-of-the-art techniques.
\end{itemize}

This paper is an extension to our conference version in \cite{zhang2021prototype}. Compared to the conference paper, this version additionally presents (\romannumeral1) a more powerful prototype completion framework for FSL, which introduces a novel part/attribute transfer network for incorporating unseen parts/attributes and develops two new methods (the EM-based and the improve EM-based methods) to estimate fusion parameters for Gaussian-based prototype fusion strategy, and improves the performance significantly; (\romannumeral2) a unified perspective to understand the mean-based prototype fusion strategy and a theoretical analysis on the Gaussian-based prototype fusion strategy; (\romannumeral3) more statistical analysis, ablation results, and visualization on miniImagenet, tieredImageNet, and CUB-200-2011, and comparisons with more state-of-the-art methods in both transductive and inductive FSL settings.

The rest of this work is organized as follows: In Section~\ref{section_2}, we briefly review related works on few-shot learning, zero-shot learning, and visual attributes. Section~\ref{section_3} describes our method in details, including the prototype completion-based meta-learning framework and the three key components, \emph{i.e.}, the part/attribute transfer network, prototype completion network and prototype fusion strategy. Section~\ref{section_4} presents and analyzes the experimental results on miniImagenet, tieredImageNet, and CUB-200-2011 data sets. Finally, the conclusion is summarized in Section~\ref{section_5}.

\section{Related Work}\label{section_2}

The key idea of the proposed prototype completion-based meta-learning framework is utilizing primitive knowledge to learn to complete prototypes for FSL. Here, the primitive knowledge refers to class-level part or attribute annotations, which can be regarded as external knowledge. Thus, in this section, many relevant studies, including few-shot learning, zero-shot learning, and visual attributes techniques, are reviewed individually.

\subsection{Few-Shot Learning}
In the literature, existing FSL methods can be divided into two groups in terms of their settings, namely the inductive FSL and transductive FSL techniques. 

\subsubsection{Inductive FSL} Most existing studies primarily address the FSL problem using the idea of inductive learning, which assumes the information of test samples is not available when performing few-shot classification tasks. Specifically, these approaches can be grouped into three categories. 1) {\bf Metric-based approaches}. The type of methods aim to learn a good metric space, where novel class samples can be nicely categorized via a nearest neighbor classifier with Euclidean \cite{snell2017prototypical}, cosine distance \cite{ChenLKWH19}, mahalanobis distance \cite{BateniGMWS20}, earth mover's distance \cite{ZhangCLS20}, or learnable distance \cite{SungYZXTH18, LiDMMHH19}. For example, Chen \emph{et al.} \cite{ChenZWC20} proposed a variational method to learn a proper scaling parameter for the Euclidean or cosine based metric, aiming to better fit the metric space to a given data distribution. 
2) {\bf Optimization-based approaches}. The methods follow the idea of modeling an optimization process over few labeled samples under the meta-learning framework, aiming to adapt to novel tasks by a few optimization steps, such as \cite{lee2019meta, sun2020meta, RajeswaranFKL19, RaghuRBV20, JamalQ19, FlennerhagRPVYH20}. 3) {\bf Semantics-based approaches}. This line of methods employ the semantic knowledge to enhance the performance of meta-learning on FSL \cite{ChenFZJXS19, boostingfew, chen2020knowledge}. For example, in \cite{zhimao2019few, babysteps2019, xing2019adaptive}, they explored the class correlations, respectively, from the perspectives of the class name, description, and knowledge graph as textual semantic knowledge, aiming to enhance the FSL classifier by the convex combination of visual and semantic modalities. Different from these works, we introduce fine-grained attributes as priors to enable a meta-learner to learn to complete prototypes for FSL, instead of to combine two modalities.

Recently, some studies turn to pre-training techniques for the FSL problem and achieve promising performance \cite{DasL20}. Chen \emph{et al.} \cite{ChenLKWH19} first proposed and investigated the pre-training techniques in FSL, by considering linear-based and cosine distance-based classifiers, respectively. In \cite{chen2020new}, a novel metric-based meta-learning method was developed by incorporating a pre-training phase. These methods, albeit delivering promising performance, do not fully explore the power of pre-training, as results show that the major improvements are made by the pre-training while the meta-learning phase contributes very marginally. According to our analysis, this is because novel classes group loosely in the pre-trained feature space. In such case, estimating more accurate and representative prototypes is more important than fine-tuning the projection spaces. Hence, in this paper, we propose a prototype completion framework to address the issue. Recently, there are also other latest pre-training FSL methods such as \cite{LiuCLL0LH20, AfrasiyabiLG20, yangfree2021, Rizve_2021_CVPR}, which focus on developing either a better pre-training strategy or a more powerful parametric classifier. Their strategies are different from our prototype completion framework.

\subsubsection{Transductive FSL} Different from inductive FSL, transductive FSL assumes that all informtation from test samples can be used for recognizing novel classes. Such approaches have been proved to be more effective than inductive FSL approaches in data-scarce scenario \cite{Qiao000HW19, DoerschGZ20, HouCMSC19}. These approaches can be divided into two groups. 1) {\bf Graph-based approaches}. The type of methods learn how to construct a good graph structure and an effective propagation mechanism from base classes as meta-knowledge, and then apply the meta-knowledge on novel classes \cite{liu2019learning, satorras2018few, KimKKY19, Tang_2021_CVPR, YangLZZZL20, RodriguezLDL20}. For instance, Yang \emph{et al.} \cite{YangLZZZL20} proposed a distribution propagation graph network for transductive FSL, aiming to propagate labels from labeled samples to unlabeled samples with the graph. 
2) {\bf Pre-training based approaches}. The methods also focus on the pre-training feature extractor and attempt to learn a classifier (e.g., SVM) \cite{DhillonCRS20, Boudiaf20, ZikoDGA20, WangXLZF20, wang2021trust, Hong_2021_CVPR} or enhance prototypes by leveraging unlabeled samples \cite{YaohuiWang_pr}. For example, Liu \emph{et al.} \cite{YaohuiWang_pr} developed a label propagation and feature shifting strategy to diminish the intra-class and cross-class prototypes bias in the pre-trained feature space. Different from these studies, we leverage the unlabeled samples to estimate prototype distribution and then leverage it to fuse prototypes. As far as we know, this is the first work to explore unlabeled samples for prototype fusion in FSL.

\subsection{Zero-Shot Learning}
Zero-shot learning (ZSL) is also closely related to FSL, which aims to address the novel class categorizations without any labeled samples \cite{guan2020zero}. The key idea is to learn a mapping function between the semantic and the visual space on the base classes, then apply the mapping to categorize novel classes. The semantic spaces in ZSL are typically attribute-based \cite{WanCLYZY019}, text description-based \cite{ReedALS16}, and word vector-based \cite{FromeCSBDRM13}. For example, in \cite{WanCLYZY019}, the semantic attributes are employed and a structure constraint on visual centers is incorporated for the mapping function learning. Our method differs from these models in two key points: (\romannumeral1) our method is for the FSL problem, where few labeled samples should be effectively utilized; (\romannumeral2) based on semantic attributes, we propose a novel prototype completion based meta-learning framework, instead of directly learning the map function. 

\begin{figure*}[!t]
	\centering
	\includegraphics[width=1.0\textwidth]{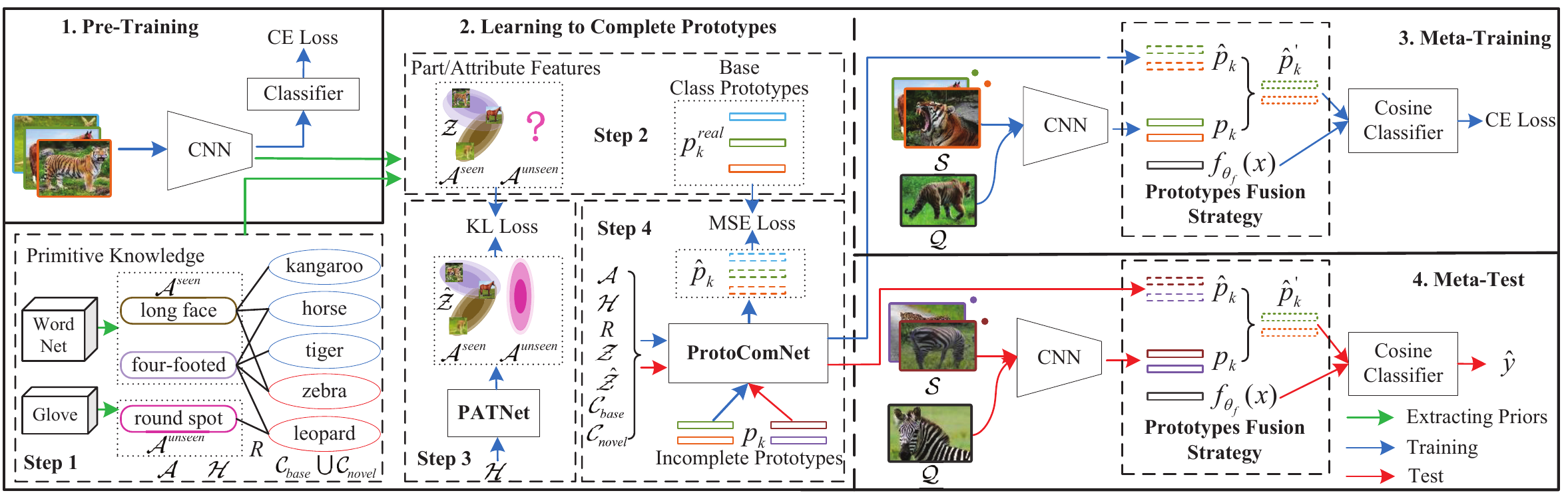} 
	\caption{The prototype completion based meta-learning framework, including four phases: (1) Pre-Training phase that learns a feature extractor by using all base classes (Section~\ref{section_3_2_1}); (2) Learning to Complete Prototypes phase that constructs primitive knowledge, extracts base class prototypes and part/attribute distribution for seen attributes, tansfers part/attributes distribution from seen parts/attributes to unseen parts/attributes, and then trains the ProtoComNet to complete prototypes (Section~\ref{section_3_2_2}); (3) Meta-Training phase that jointly fine-tunes the feature extractor and ProtoComNet in an episodic training manner (Section~\ref{section_3_2_3}); and (4) Meta-Test phase that performs novel class prediction (Section~\ref{section_3_2_4}).}
	\label{fig2}
\end{figure*}

\subsection{Visual Attributes}
Visual attributes refer to the visual features of object components \cite{multilabelobjectattribute}, which have been successfully utilized in various domains, such as action recognition \cite{ZhangTGL18}, zero-shot learning \cite{WanCLYZY019, XianLSA19}, person Re-ID \cite{LinZZWHYY19}, and image caption \cite{ChenDLZH18}. Recently, several FSL techniques relying on visual attributes have been proposed. In \cite{tokmakov2019learning}, an attribute decoupling regularizer was developed based on visual attributes to obtain good representations for images. 
Hu \emph{et al.} \cite{Hu2019Weaklys} proposed a compositional feature aggregation module to explore both spatial and semantic visual attributes for FSL. Zou \emph{et al.} \cite{zou2020compositional} explored compositional few-shot recognition by learning a feature representation composed of important visual attributes. All the methods utilize visual attributes for better representations. Different from these studies, we leverage them to learn a prototype completion strategy. As a result, more accurate prototypes can be obtained for FSL. 

\section{Methodology}
\label{section_3}
In this section, we first present a formal definition of the FSL problem setting. Second, the proposed prototype completion based meta-learning framework is introduced. Finally, the three key components in the framework, namely the parts/attribute transfer network, the prototype completion network, and the prototype fusion strategy are elaborated in the last three subsections, respectively. 

\subsection{Problem Definition}
For $N$-way $K$-shot FSL problems, we are given two sets: a training set $\mathcal{S}=\{(x_i, y_i)\}_{i=0}^{N \times K}$ with a few of labeled samples  (called support set) and a test set $\mathcal{Q}=\{(x_i, y_i)\}_{i=0}^{M}$ consisting of unlabeled samples (called query set). Here $x_i$ denotes the image sampled from the set of novel classes $\mathcal{C}_{novel}$, $y_i \in \mathcal{C}_{novel}$ is the label of $x_i$, $N$ indicates the number of classes in $\mathcal{S}$, $K$ denotes the number of images of each class in $\mathcal{S}$, and $M$ denotes the number of images in $\mathcal{Q}$. Meanwhile, we also have an auxiliary data set with abundant labeled images $\mathcal{D}_{base}=\{(x_i, y_i)\}_{i=0}^{B}$, where $B$ is the number of images in $\mathcal{D}_{base}$, the image $x_i$ is sampled from the set of base classes $\mathcal{C}_{base}$, \emph{i.e.}\ $y_i \in \mathcal{C}_{base}$, and the sets of class $\mathcal{C}_{base}$ and $\mathcal{C}_{novel}$ are disjoint. Our goal is to learn a classifier for the query set $\mathcal{Q}$ on the support set $\mathcal{S}$ and the auxiliary data set $\mathcal{D}_{base}$. We note that the query set $\mathcal{Q}$ is available by regarding it as a set of unlabeled samples to transductive FSL. However, it is not accessible for inductive FSL.

\subsection{Overall Framework}
\label{section_3_2}
As shown in Fig.~\ref{fig2}, the proposed prototype completion-based meta-learning framework consists of four phases, including pre-training, learning to complete prototypes, meta-training, and meta-test. Next, we detail them respectively. 

\subsubsection{Pre-Training} 
\label{section_3_2_1}
In this phase, following \cite{ChenLKWH19, chen2020new, RodriguezLDL20}, we build and train a convolution neural network (CNN) classifier with the base classes. 
Then, the last softmax layer is removed and the classifier turns into a feature extractor $f_{\theta_f}()$. This produces a good embedding representation for each image. 


\subsubsection {Learning to Complete Prototypes} 
\label{section_3_2_2}
We propose a Prototype Completion Network (ProtoComNet) as a meta-learner. It accounts for complementing the missing attributes for incomplete prototypes. The main details of the ProtoComNet will be elaborated in Section~\ref{section3_4}. Here we first give an overview of its workflow depicted in Fig.~\ref{fig2}, which includes four steps: 

{\bf Step 1}. We construct primitive knowledge for all classes. The knowledge is what kinds of attribute feature the class should have, \emph{e.g.}, the leopard has four feet and round spot, and zebra has long face and four feet. We note that such kinds of knowledge is very cheap to obtain, \emph{e.g.}, from WordNet. 
Let $\mathcal{A} = \{a_i\}_{i=0}^{F-1}$ denotes the set of class parts/attributes where $F$ is the number of attributes, and $R$ denotes the association matrix between the attributes and the classes, where $R_{ka_i}=1$ if the attribute $a_i$ is associated with the class $k$; otherwise $R_{ka_i}=0$. Meanwhile, the semantic embeddings of all classes and attributes are calculated by Glove \cite{pennington2014glove} in an average manner of word embeddings, denoted by $\mathcal{H} =\{h_k\}_{k=0}^{|\mathcal{C}_{base}|+|\mathcal{C}_{novel}|-1} \cup \{h_{a_i}\}_{i=0}^{F-1}$. In particular, we split the set of class parts/attributes $\mathcal{A}$ into two subset: $\mathcal{A}^{seen}$ and $\mathcal{A}^{unseen}$ (i.e., $F=|\mathcal{A}^{seen}|+|\mathcal{A}^{unseen}|$). The former $\mathcal{A}^{seen}$ denotes the set of parts/attributes that base classes contains. On the other hand, the latter $\mathcal{A}^{unseen}$ refers to the set of parts/attributes that the novel classes contain but does not appear in base classes.

{\bf Step 2}. Based on the pre-trained feature extractor $f_{\theta_f}()$ and the above primitive knowledge, we extract two types of information as priors, namely base class prototypes and seen part/attribute features. Specifically, the base class prototypes $p_k^{real}$ can be calculated by averaging the extracted features of all samples in the base class $k$, that is,
\begin{equation} 
p_k^{real}=\frac {1}{|\mathcal{D}_{base}^k|} \sum_{(x, y) \in \mathcal{D}_{base}^k} f_{\theta_f}(x),
\label{eq1}
\end{equation}
where $\mathcal{D}_{base}^k$ denotes the set of samples from the base class $k$. As for the feature $z_{a_i}$ of each seen part/attribute $a_i \in \mathcal{A}^{seen}$, we denote all base class samples that have the corresponding part/attribute $a_i \in \mathcal{A}^{seen}$ in the primitive knowledge as a set $D_{base}^{a_i}$. Then, we calculate its mean $\mu_{a_i}$ and diagonal covariance $diag(\sigma_{a_i}^2)$ as: 
\begin{equation} 
\mu_{a_i}=\frac {1}{|\mathcal{D}_{base}^{a_i}|} \sum_{(x, y) \in \mathcal{D}_{base}^{a_i}} f_{\theta_f}(x),
\label{eq2}
\end{equation}
\begin{equation} 
\sigma_{a_i}=\sqrt{\frac {1}{|\mathcal{D}_{base}^{a_i}|} \sum_{(x, y) \in \mathcal{D}_{base}^{a_i}} (f_{\theta_f}(x)\ -\ \mu_{a_i})^2}.
\label{eq3}
\end{equation}
Here, the mean $u_{a_i}$ and the diagonal covariance $diag(\sigma_{a_i}^2)$ characterize the part/attribute feature distribution of each seen part/attribute $a_i \in \mathcal{A}^{seen}$, \emph{i.e.}, $z_{a_i} \sim N(\mu_{a_i}, diag(\sigma_{a_i}^2))$, which will be used in Section~\ref{section3_3} and \ref{section3_4}.

{\bf Step 3}. According to Eqs.~\ref{eq2} and~\ref{eq3}, we can estimate the feature distribution of the seen parts/attributes $a_i \in \mathcal{A}^{seen}$. However, the method fails to model the unseen parts/attributes  $a_i \in \mathcal{A}^{unseen}$ since it does not appear in base classes. To address the drawback, we design a \underline{P}art/\underline{A}ttribute \underline{T}ransfer \underline{Net}work (PATNet) $f_{\theta_p}()$ with parameters $\theta_p$, which accounts for inferring the feature distribution of unseen parts/attributes by exploring the semantics relationship between unseen and seen parts/attributes. The intuition behind it is that the similar parts/attributes in semantics should have a similar feature distribution. Its design details will be introduced in Section~\ref{section3_3}. Here, we focus on introducing the overall workflow of the PATNet. Specifically, we take the semantic embedding $\{h_{a_i}\}_{i=0}^{|\mathcal{A}^{seen}|-1}$ of all seen parts/attributes $\{a_i\} \in \mathcal{A}^{seen}$ as inputs, and treat the feature distribution $N(\mu_{a_i}, diag(\sigma_{a_i}^2))$ of the seen parts/attributes $a_i \in \mathcal{A}^{seen}$ estimated by Eqs.~\ref{eq2} and~\ref{eq3} as prediction targets, to train the proposed PATNet $f_{\theta_p}()$ by using the Kullback-Leibler (KL) divergence loss. That is,

\begin{small}
	\begin{equation} 
	\begin{aligned}
	\hat{\mu}_{a_i},\ &\hat{\sigma}_{a_i} = f_{\theta_p}(h_{a_i}),\ i=0,1,...,|\mathcal{A}^{seen}|-1\\
	\min\limits_{\theta_p} \mathbb{E}_{a_i \in \mathcal{A}^{seen}} \ KL(&N(\hat{\mu}_{a_i}, diag(\hat{\sigma}_{a_i}^2)),\ N(\mu_{a_i}, diag(\sigma_{a_i}^2))),
	\label{eq4}
	\end{aligned}
	\end{equation}
\end{small}where $KL()$ denotes the Kullback-Leibler (KL) divergence loss. Then, we train the parts/attributes transfer network $f_{\theta_p}()$ until it converges. The well trained PATNet can infer the feature distribution of each seen and unseen part/attribute through its semantics. As a result, we obtain a new feature distribution $\hat{z}_{a_i} \sim N(\hat{\mu}_{a_i}, diag(\hat{\sigma}_{a_i}^2))$ for each seen and unseen parts/attribute by utilizing its semantics as input of PATNet, which will be used in Section~\ref{section3_4}.

{\bf Step 4}. Upon the results of the previous steps, we mimic the setting of $K$-shot tasks and construct a set of prototype completion tasks to train our meta-learner $f_{\theta_c}()$ (\emph{i.e.}, ProtoComNet) in an episodic manner \cite{vinyals2016matching}. Specifically, in each episode, we first randomly select one class $k$ from base classes $C_{base}$ and $K$ images for the class $k$ from $\mathcal{D}_{base}$ as support set $S$. Then, we average the features of all samples in $S$ as the incomplete prototypes $p_k$. Here, we consider it as incomplete because some representative features may be missing. Even though in some cases this may not be true, regarding them as incomplete ones does no harms to our meta-learner. Finally, we take the incomplete prototypes $p_k$, the primitive knowledge (the class-attribute association matrix $R$ and word embedding $\mathcal{H}$), and the parts/attributes features $\mathcal{Z}=\{z_{a_i}\}_{i=0}^{|\mathcal{A}^{seen}|-1}$ and $\mathcal{\hat{Z}}=\{\hat{z}_{a_i}\}_{i=0}^{|\mathcal{A}^{seen}|+|\mathcal{A}^{unseen}|-1}$ as inputs, and treat the base class prototypes $p_k^{real}$ as outputs, to train our meta-learner by using the Mean-Square Error (MSE) loss. That is,
\begin{equation} 
\min\limits_{\theta_c} \mathbb{E}_{(p_k,\ p_k^{real}) \in \mathbb{T}} \ MSE(f_{\theta_c}(p_k, R, \mathcal{H}, \mathcal{Z}, \mathcal{\hat{Z}}),\ p_k^{real}),
\label{eq5}
\end{equation}
where $\theta_c$ denotes the parameters of our meta-learner and $\mathbb{T}$ denotes the set of prototype completion tasks.

\subsubsection{Meta-Training} 
\label{section_3_2_3}
To jointly fine-tune the feature extractor $f_{\theta_f}()$ and the meta-learner $f_{\theta_c}()$, we construct a number of $N$-way $K$-shot tasks from $\mathcal{D}_{base}$ following the episodic training manner \cite{vinyals2016matching}. Specifically, in each episode, we sample $N$ classes from the base classes $\mathcal{C}_{base}$, $K$ images in each class 
as the support set $\mathcal{S}$, and $M$ images as the query set $\mathcal{Q}$. Then, $f_{\theta_f}()$ and $f_{\theta_c}()$ can be further fine-tuned by maximizing the likelihood estimation on query set $Q$. That is,
\begin{equation}
\begin{aligned}
\max\limits_{\theta}  \mathbb{E}_{(\mathcal{S},\mathcal{Q}) \in \mathbb{T}'} \sum_{(x,y) \in \mathcal{Q}} log(P(y|x, \mathcal{S}, R, \mathcal{H}, \mathcal{Z}, \mathcal{\hat{Z}}, \theta)),
\end{aligned}
\label{eq6}
\end{equation}
where $\theta = \{\theta_f, \theta_c\}$ and $\mathbb{T}'$ denotes the set of $N$-way $K$-shot tasks. Specifically, for each episode, we first estimate its class prototype $p_k$ by averaging the features of the labeled samples. That is,
\begin{equation}
p_k = \frac{1}{|\mathcal{S}_k|} \sum_{x \in \mathcal{S}_k} f_{\theta_f}(x),
\label{eq7}
\end{equation}
where $\mathcal{S}_k$ is the support set extracted for the class $k$. 
Then, the ProtoComNet is applied to complete $p_k$, and we have: 
\begin{equation}
\hat{p}_k = f_{\theta_c}(p_k, R, \mathcal{H}, \mathcal{Z}, \mathcal{\hat{Z}})).
\label{eq8}
\end{equation}
Moreover, to obtain more reliable prototypes, we further explore unlabeled samples and combine $p_k$ and $\hat{p}_k$ by introducing a Gaussian-based prototype fusion strategy (which will be introduced in Section~\ref{section3_5}). As a result, the fused prototype $\hat{p}'_k$ is obtained. 
Finally, the probability of each sample $x \in \mathcal{Q}$ to be class $k$ is estimated based on the proximity between its feature $f_{\theta_f}(x)$ and $\hat{p}'_k$. That is,
\begin{equation}
P(y=k|x, \mathcal{S}, R, \mathcal{H}, \mathcal{Z}, \theta) = \frac{e^{d(f_{\theta_f}(x),\ \hat{p}'_k)\ \cdot\ \gamma}}
{\sum_{c} e^{d(f_{\theta_f}(x),\ \hat{p}'_c)\ \cdot\ \gamma}},
\label{eq9}
\end{equation}
where $d()$ denotes the cosine similarity of two vectors and $\gamma$ is a learnable scale parameter.

\subsubsection{Meta-Test} 
\label{section_3_2_4}
Following Eqs. (\ref{eq7}) $\sim$ (\ref{eq9}), we directly perform few-shot classification for novel class prediction.

\subsection{Part/attribute Transfer Network}
\label{section3_3}
In this subsection, we introduce the first key component of learning to complete prototypes (Step 3 in Section~\ref{section_3_2_2}), namely the PATNet $f_{\theta_p}()$. Our intuition is that the similar parts/attributes in semantics should have a similar feature distribution. Thus, we directly treat the semantic embeddings $\{h_{a_i}\}_{i=0}^{F-1}$ of part/attribute as input and the parts/attributes distribution $N(\mu_{a_i}, diag(\sigma_{a_i}^2))$ as output to build the PATNet. 

\begin{figure}[!t]
	\centering
	\includegraphics[width=0.95\columnwidth]{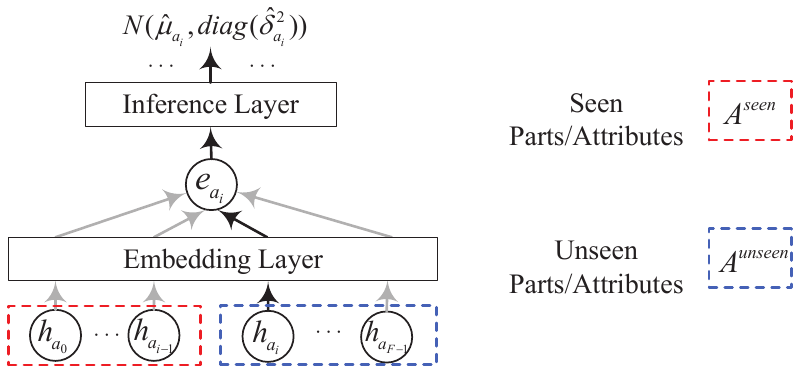} 
	\caption{Illustration of the proposed parts/attributes transfer networks. 
	}
	\label{fig33}
\end{figure}

As shown in Fig.~\ref{fig33}, the network consists of an embedding layer $f_{\theta_{pe}}()$ and an inference layer $f_{\theta_{pi}}()$, where $\theta_{pe}$ and $\theta_{pi}$ denote their parameters, respectively. Here, the former aims to map each semantic embeddings to a new embedding space, and then the latter accounts for estimating the feature distribution of each part/attribute. Next, we detail them, respectively.

{\bf Embedding Layer.} We take the semantic embedding $h_{a_i}$ of each part/attribute as input of the embedding layer $f_{\theta_{pe}}()$, and then project the semantic embedding $h_{a_i}$ to a new embedding space. As a result, the new embedding $h'_{a_i}$ can be obtained. That is,
\begin{equation} 
h'_{a_i}\ =\ f_{\theta_{pe}}(h_{a_i}).
\end{equation}

{\bf Inference Layer.} Based on the new embedding $h'_{a_i}$, we employ an inference layer consisting of a mean module and a diagonal covariance module to predict the distribution of each seen and unseen part/attribute, which is characterized by a multivariate normal distribution parameterized with its mean $\hat{\mu}_{a_i}$ and diagonal convariance $diag(\hat{\sigma}_{a_i}^2)$. That is,
\begin{equation} 
\begin{aligned}
\hat{\mu}_{a_i}, \hat{\sigma}_{a_i} &= f_{\theta_{pi}}(h'_{a_i}), \\
\hat{z}_{a_i} \ \sim \ N(&\hat{\mu}_{a_i}, \ diag(\hat{\sigma}_{a_i}^2)).
\end{aligned}
\end{equation}
Note that $\theta_{p}$ contains the two parameters $\theta_{pe}$ and $\theta_{pi}$.

\subsection{Prototype Completion Network}
\label{section3_4}
\begin{figure}[t]
	\centering
	\includegraphics[width=0.95\columnwidth]{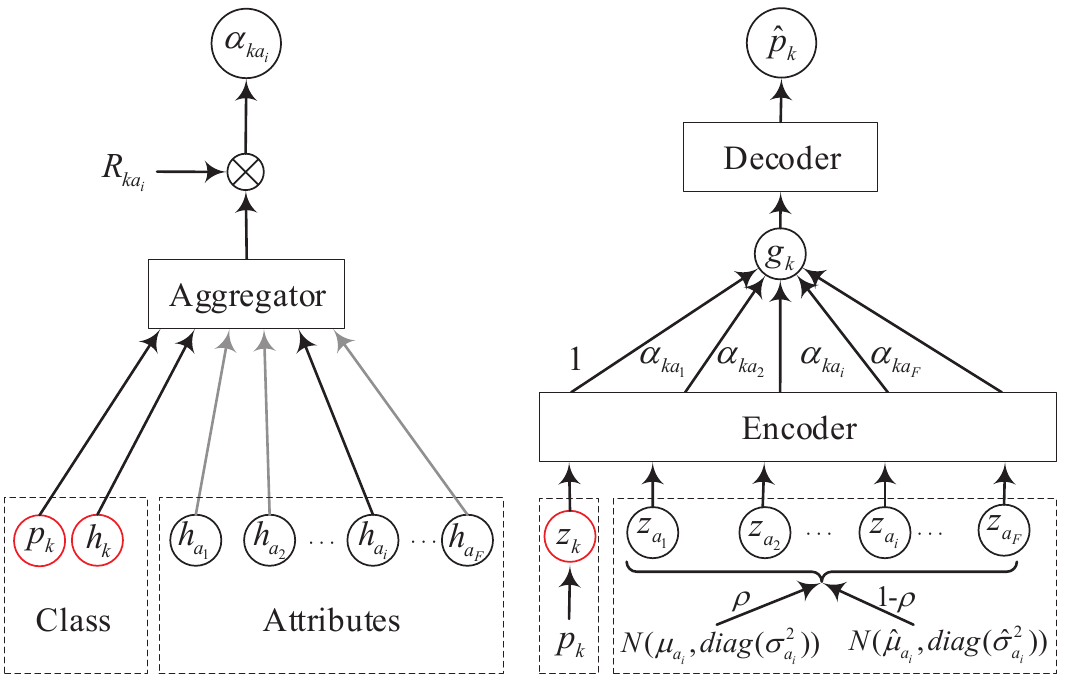} 
	\caption{Illustration of the encoder-aggregator-decoder networks. 
	}
	\label{fig3}
\end{figure}

In this subsection, we introduce how the ProtoComNet $f_{\theta_c}()$ are designed, which is the second key component for learning to complete prototypes (Step 4 in Section~\ref{section_3_2_2}). Our intuition is that the parts/atributes feature can be transfered from base classes to novel classes for prototype completion. For example, even if human haven't seen ``zebra'', they can also imagine its visual features of ``long face'' once they learn this knowledge from ``kangaroo'' and ``horse''. Thus, we treat the primitive knowledge ($\cal R$ and $\cal H$), part/attribute features $\cal Z$ and $\mathcal{\hat{Z}}$ and the incomplete prototypes $p_k$ as input and the completed prototypes $\hat{p}_k$ as output, and then build an encoder-aggregator-decoder network, as shown in Fig.~\ref{fig3}. Here, the encoder aims to form a low-dimensional representation for prototypes and part/attributes. Then, the aggregator accounts for evaluating the importance of different parts/attributes and combining them with a weighted sum. Finally, the decoder is in charge of the prediction of complete prototypes ${\hat p}_k$. Next, we detail them, respectively.

{\bf The Encoder.} 
In the training part, the encoding process involves a sampling step of an attribute feature $z_{a_i}$ from its distribution $N(\mu_{a_i}, diag(\sigma_{a_i}^2))$ or $N(\hat{\mu}_{a_i}, diag(\hat{\sigma}_{a_i}^2))$, followed by an encoder $g_{\theta_{ce}}()$ that encodes the attribute feature $z_{a_i}$ and the estimated prototypes $p_k$ to a latent code $z'_{a_i}$ and $z'_k$, respectively. To enhance the generalization of the model for seen and unseen parts/attributes, we adopt a randomized manner with a probability $\rho=0.5$ to sample the attribute feature $z_{a_i}$ from seen part/attribute distribution $N(\mu_{a_i}, diag(\sigma_{a_i}^2))$ and unseen part/attribute distribution $N(\hat{\mu}_{a_i}, diag(\hat{\sigma}_{a_i}^2))$. The overall encoding process is formally expressed as:
\begin{equation} 
\begin{aligned}
z_{a_i} \ &\sim 
\begin{cases}
N(\mu_{a_i}, \ diag(\sigma_{a_i}^2)), & a_i \in \mathcal{A}^{seen} \&\  r<\rho\\
N(\hat{\mu}_{a_i}, \ diag(\hat{\sigma}_{a_i}^2)), & otherwise
\end{cases}, \\&
\ z_{a_i}' = g_{\theta_{ce}}(z_{a_i}), \\&
z_{k} \ = p_k
,\ z_{k}' = g_{\theta_{ce}}(z_{k}),
\end{aligned}
\label{eq10}
\end{equation}
where $\theta_{ce}$ denotes the parameters of the encoder and $r$ is a random number from 0 to 1. Note that, in the meta-test phase, we regard $N(\mu_{a_i}, diag(\sigma_{a_i}^2))$ as the feature distribution of seen parts/attributes and $N(\hat{\mu}_{a_i}, diag(\hat{\sigma}_{a_i}^2))$ as the ones of unseen parts/attributes; and we remove the sampling step and use the mean $\mu_{a_i}$ and $\hat{\mu}_{a_i}$ to replace $z_{a_i}$.

{\bf The Aggregator.} Intuitively, different parts/attributes make varying contributions to distinct classes, for example, the ``nose'' is more representive for elephants than tigers to complete their prototypes. Hence, differentiating their contributions in the completion is important. To this end, we employ an attention-based aggregator $g_{\theta_{ca}}()$. Here, we calculate the attention weights $\alpha_{ka_i}$ by using the semantic embeddings $h_{k}$ and $h_{a_i}$ of the class $k$ and the attribute $a_i$, and the incomplete prototypes $p_k$. Then, we apply them to combine the latent codes $z_{k}'$ and $z_{a_i}'$, and obtain the aggregated result $g_k$ as follows: 

\begin{equation} 
\begin{aligned}
\alpha_{ka_i}&=R_{ka_i}g_{\theta_{ca}}(p_{k} || h_{k} || h_{a_i})
, \\ g_k &= \sum_{a_i}\alpha_{ka_i}z'_{a_i}+z'_k ,
\end{aligned}
\label{eq11}
\end{equation}
where $\theta_{ca}$ is the parameters of the aggregator and $||$ is a concatenation operation.

{\bf The Decoder.} Finally, we use the aggregated result $g_k$ to decode the complete prototypes $\hat{p}_{k}$ for each class $k$ by the decoder module $g_{\theta_{cd}}()$. That is, 
\begin{equation} 
\begin{aligned}
\hat{p}_{k} = g_{\theta_{cd}}(g_k),
\end{aligned}
\end{equation}
where $\theta_{cd}$ denotes the parameters of the decoder. 

\subsection{Prototype Fusion Strategy}
\label{section3_5}
Till now, we have two prototype estimations, \emph{i.e.}, the mean-based prototypes $p_k$ and the completed prototypes ${\hat p}_k$. Next, we will discuss why and how to fuse these two estimations from the perspective of Bayesian estimation.

\subsubsection {Why do we fuse prototypes?} 
\begin{figure}[!t]
	\centering
	\subfigure[Experiment on 5-way $K$-shot task]{ 
		\label{fig3_4_a} 
		\includegraphics[width=0.51\columnwidth]{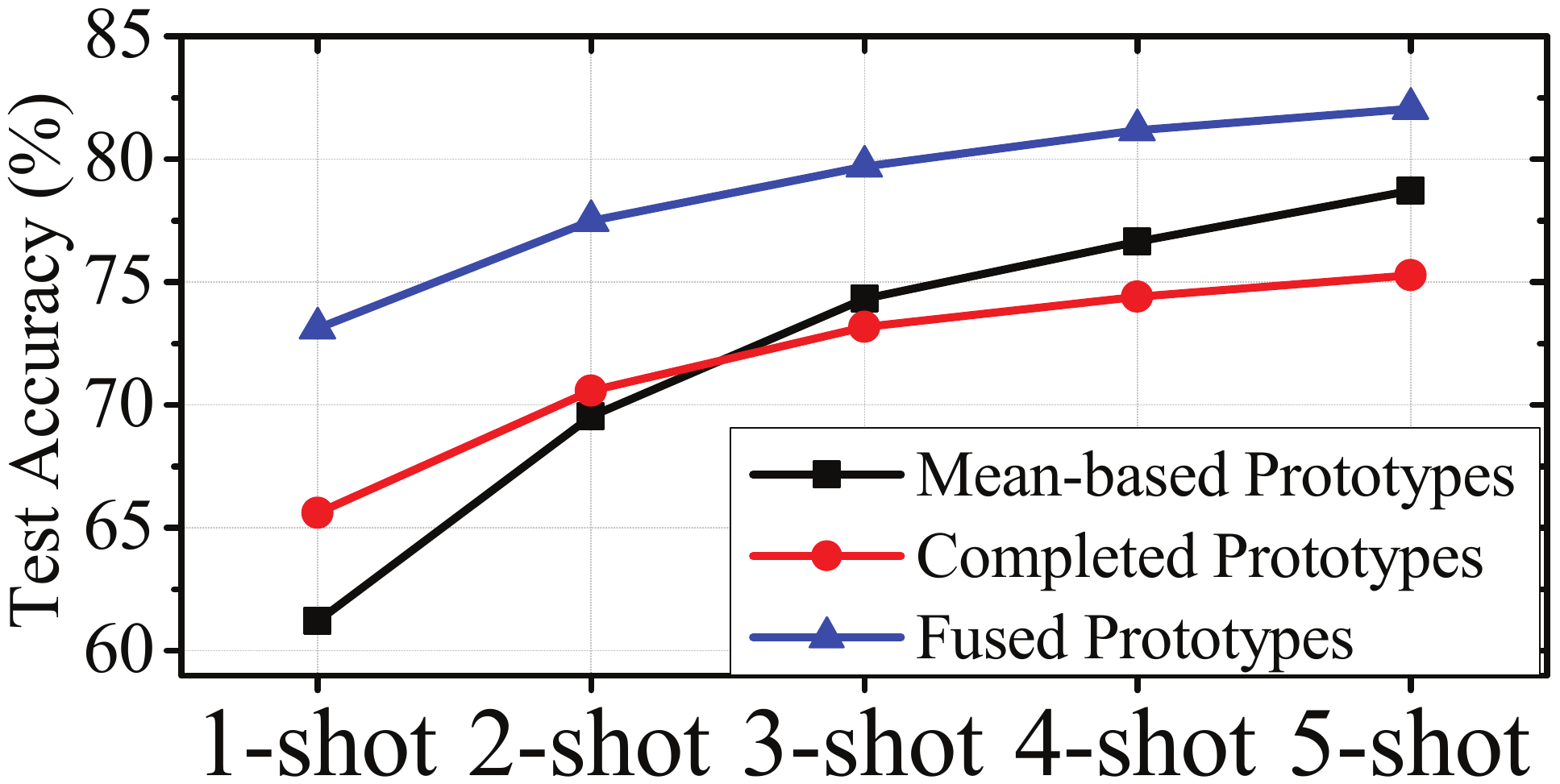}}
	\subfigure[Prototype fusion strategy]{ 
		\label{fig3_4_b} 
		\includegraphics[width=0.45\columnwidth]{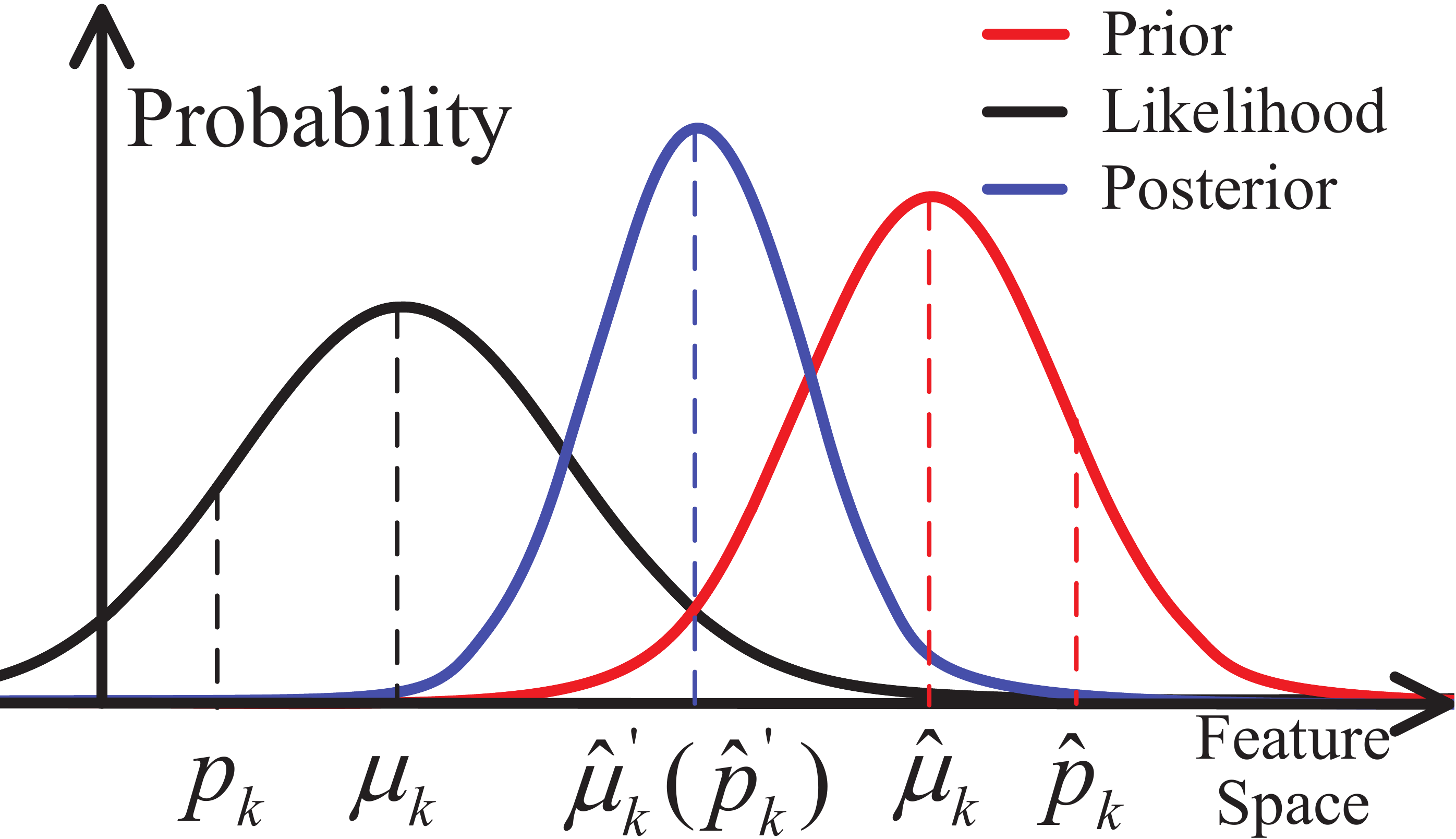}}
	\caption{Test accuracy of $p_k$ and ${\hat p}_k$ on 5-way $K$-shot tasks of miniImagenet (a) and illustration of prototype fusion strategy (b). }
	\label{fig3_4}
\end{figure}

Actually, both the estimates $p_k$ and ${\hat p}_k$ have their own biases. The former is mainly due to the scarcity or incompleteness of labeled samples in novel classes, which produces biased means; while the latter is brought by the primitive knowledge noises and the base-novel class differences. The fact implies that the two estimates can remedy each other. When the labeled samples are very scarce and incomplete, the completed prototypes ${\hat p}_k$ are more reliable because the completion is learned from a great number of base class tasks. As more and more labeled samples become available, the mean-based prototypes are more representative because the ProtoComNet may result in prototype completion error problem under the effects of primitive knowledge noises or class differences. Fig.~\ref{fig3_4_a} shows an example to demonstrate this. We observe that the completed prototypes are more accurate on 1/2-shot tasks while the mean-based ones are better on 3/4/5-shot tasks. Thus, a prototype fusion strategy is desired to combine their advantages and form more representative prototypes. 

\subsubsection{How to fuse prototypes?} 
\label{section_3_5_2}
We apply the Bayesian estimation to fuse the two kinds of prototypes. Specifically, we assume that the estimated prototypes follow a Multivariate Gaussian Distribution (MGD), as the samples in the pre-trained space are continuous and clustered together (shown in Fig.~\ref{fig1}). Based on this assumption, $p_k$ can be regarded as a sample from the MGD with mean $\mu_k$ and diagonal covariance $diag(\sigma_{k}^2)$, \emph{i.e.}, $N(\mu_k, diag(\sigma_{k}^2))$. Likewise, ${\hat p}_k$ is a sample from $N({\hat \mu}_k, diag(\hat{\sigma}_{k}^2))$ with mean ${\hat \mu}_k$ and diagonal covariance $diag(\hat{\sigma}_{k}^2)$. As shown in Fig.~\ref{fig3_4_b}, from the view of Bayesian estimation, we regard the distribution $N({\hat \mu}_k, diag(\hat{\sigma}_{k}^2))$ as a prior, and treat the distribution $N(\mu_k, diag(\sigma_{k}^2))$ as the conditional likelihood of observed few labeled samples. Then, the Beyesian estimation of fused prototypes can be expressed as their product, \emph{i.e.}, a posterior MGD $N({\hat \mu}'_k, diag(\hat{\sigma}_{k}'^2))$ with mean ${\hat \mu}'_k=\frac{\sigma_{k}^2 \odot \hat{\mu}_{k} + \hat{\sigma}_{k}^2 \odot \mu_{k}}{\hat{\sigma}_{k}^2 + \sigma_{k}^2}$ and diagonal covariance $diag(\hat{\sigma}_{k}'^2)=diag(\frac{\sigma_{k}^2 \odot \hat{\sigma}_{k}^2}{\hat{\sigma}_{k}^2 + \sigma_{k}^2})$, where $\odot$ is element-wise product (Please refer to Appendix~\ref{appendix_a} for its derivations). Finally, we take the mean $\hat{\mu}'_{k}$ as the fused prototypes $\hat{p}'_{k}$ to solve the few-shot tasks (Please refer to Section~\ref{section_3_5_5} for its theoretic analysis). 

In this paper, we term the overall Bayesian estimation procedure as Gaussian-based prototype fusion strategy (GaussFusion). We can see that ${\hat \mu}'_k$ is determined by four unknown variables $\mu_{k}$, $\sigma_{k}$, $\mu'_{k}$, and $\sigma'_{k}$. Next, we introduce four types of methods to estimate them.

\subsubsection{How to estimate $\mu_{k}$, $\sigma_{k}$, $\mu'_{k}$, and $\sigma'_{k}$?}
\label{section_3_5_3}
In this part, we discuss four methods to estimate the four unknown variables $\mu_{k}$, $\sigma_{k}$, $\mu'_{k}$, and $\sigma'_{k}$, including (\romannumeral1) assumption-based estimation method, (\romannumeral2) two-step estimation method, (\romannumeral3) EM-based estimation method, and (\romannumeral4) improved EM-based estimation method. Among them, the methods (\romannumeral1) and (\romannumeral2) belong to non-iterative approaches, where the former follows the estimate strategy proposed in \cite{XueW20} and the latter is our conference strategy \cite{zhang2021prototype}. The rest of these methods (i.e., the methods (\romannumeral3) and (\romannumeral4)) all are iterative approaches, which are newly-developed in this paper.

{\bf Assumption-based Estimation method.} The Mean-based Prototype Fusion (MeanFusion) strategy proposed in \cite{XueW20} regards the averaged prototypes as the fused prototypes ${\hat p}'_k = 0.5(p_{k} + \hat{p}_{k})$. This strategy can be considered as a special case of our GaussFusion, where we assume that the two means satify $\mu_{k}=p_k$ and ${\hat \mu}_k = {\hat p}_k$, and the two diagonal covariance is also equal, i.e., $\sigma_{k} = \hat{\sigma}_{k}$. However, the assumption is too strong to fit the real prototype distribution. Thus, the performance improvement of the MeanFusion is limited for FSL.

{\bf Two-Step Estimation Method.} Inspired by transductive FSL \cite{YaohuiWang_pr}, we propose to estimate the four variables by leveraging the unlabeled samples in a two-step manner: {\bf Step 1)} we calculate the probability of each sample $x \in \mathcal{S} \cup \mathcal{Q} $ belonging to class $k$ by regarding $p_k$ and $\hat{p}_{k}$ as the prototypes, respectively. For example, when we take $p_k$ as the prototypes, the probability of each unlabeled sample $x \in  \cal Q $ can be computed as: 
\begin{equation}
P(y=k|x) = \frac{e^{d(f_{\theta_f}(x),\ p_k)\ \cdot\ \lambda}}
{\sum_{c} e^{d(f_{\theta_f}(x),\ {p}_c)\ \cdot\ \lambda}},
\label{eq12}
\end{equation}
where $d()$ indicates the cosine similarity of two vectors and $\lambda$ is a hyper-parameter. Following \cite{ChenLKWH19}, $\lambda=10$ is used. As for each labeled sample $x \in \mathcal{S}$, the probability turns into a one-hot vector by its labels. $\hat{P}(y=k|x)$ can be computed in a similar manner by using prototypes $\hat{p}_{k}$. {\bf Step 2)} we take $P(y=k|x)$ as sample weights and estimate the mean $\mu_{k}$ and the diagonal covariance $diag(\sigma_{k}^2)$ of each prototype distribution in a weighted average manner. That is,
\begin{small}
	\begin{equation} 
	\begin{aligned}
	\mu_{k} = \frac{1}{\sum \limits_{x \in \mathcal{S} \cup \mathcal{Q}}P(k|x)} \sum_{x \in \mathcal{S} \cup \mathcal{Q}} P(k|x) f_{\theta_f}(x),
	\end{aligned}
	\label{eq13}
	\end{equation}
\end{small}
\begin{small}
	\begin{equation} 
	\begin{aligned}
	\sigma_{k} = \sqrt{\frac{1}{\sum \limits_{x \in \mathcal{S} \cup \mathcal{Q}}P(k|x)} \sum_{x \in \mathcal{S} \cup \mathcal{Q}} P(k|x) (f_{\theta_f}(x) - \mu_{k})^2}.
	\end{aligned}
	\label{eq14}
	\end{equation}
\end{small}Similarly, the mean $\hat{\mu}_{k}$ and the diagonal covariance $diag(\hat{\sigma}_{k}^2)$ can be calculated in a similar manner by regarding $\hat{P}(y=k|x)$ as sample weights. The two step prediction strategy is the method proposed in our conference version \cite{zhang2021prototype}.

{\bf EM-based Estimation Method.} The EM (Expectation-Maximization) algorithm \cite{dempster1977maximum} is a widely used parameter estimation method, which adopts an iterative strategy to polish the parameter estimation. Thus, we attempt to estimate the above four variables by employing the EM algorithm. Specifically, we regard the support and query samples $x \in \mathcal{S} \cup \mathcal{Q}$ as the observation data from Gaussian mixture distribution with unknown mean $\mu_{k}$ or $\hat{\mu}_{k}$ and diagonal covariance $diag(\sigma_{k})$ or $diag(\hat{\sigma}_{k})$ ($k=0,1,...,N-1$), and regard the prototypes $p_k$ or ${\hat p}_k$ as the initial mean of the $k$-th Gaussian distribution. Our goal is to fit the mean and diagonal covariance to the observation data $x \in \mathcal{S} \cup \mathcal{Q}$. That is, maximizing the likelihood estimate for $\mu_{k}$ and $\sigma_{k}$ (Note that $\hat{\mu}_{k}$ and $\hat{\sigma}_{k}$ are similar) as:
\begin{equation}
l(\{\mu_{k}, \sigma_{k}\}_{k=0}^{N-1}) = \prod_{x \in \mathcal{S} \cup \mathcal{Q}} \sum_{k=0}^{N-1} z \cdot N(x; \mu_{k}, diag(\sigma_{k}^2)),
\label{eq15}
\end{equation} where $z$ is a hidden variable denoting the posterior probability that $x$ belongs to class $k$.

We adopt EM algorithm to optimize Eq.~\ref{eq15}, which includes following three steps: 1) initializing the mean $\mu_{k}$ or $\hat{\mu}_{k}$ by using the prototypes $p_k$ or ${\hat p}_k$ and diagonal covariance $\sigma_{k}$ or $\hat{\sigma}_{k}$ in a constant (We empirically find that our method can obtain high classification peformance when it is set as 35); 2) Performing {\bf E step} to estimate the posterior probability $z$ that a given observation $x$ belongs to a given class $k$ by using the probability density function $N(x; \mu_{k}, diag(\sigma_{k}^2))$ or $N(x; \hat{\mu}_{k}, diag(\hat{\sigma}_{k}^2))$. Note that we estimate the probability of each support sample $x \in \mathcal{S}$ by a one-hot vector of its label since its label is known; 3) Performing {\bf M step} to maximize the posterior probability and find the optimal mean $\mu_{k}$ or $\hat{\mu}_{k}$ and diagonal covariance $\sigma_{k}$ or $\hat{\sigma}_{k}$; 4) Repeatedly carrying out these two steps (i.e., {\bf E step} and {\bf M step}) until convergence. Finally, we take the resulting $\mu_{k}$ or $\hat{\mu}_{k}$ and $\sigma_{k}$ or $\hat{\sigma}_{k}$ as our estimation.

{\bf Improved EM-based Estimation method.} In the above EM-based method, the posterior probability $z$ is estimated by using the Gaussian probability density. Its  calculation is similar to the Mahalanobis distance. However, recent studies \cite{ChenLKWH19, chen2020new} found that the cosine distance-based classifier show better performance on the estimation of posterior probability $z$ for FSL. Inspired by this fact, we estimate it by leveraging the cosine-based classifier (i.e., Eq.~(\ref{eq12})). In particular, the improved EM-based method can be regarded as an extension of the above Two-Step Method by using the EM algorithm. Specifically, we first initialize the mean $\mu_{k}$ or $\hat{\mu}_{k}$ by using the prototypes $p_k$ or ${\hat p}_k$. Second, the step {\bf 1)} (described in Two-Step Estimation Method) can be regarded as an E-Step, i.e., regarding the mean $\mu_{k}$ or $\hat{\mu}_{k}$ as the prototypes of cosine classifier and then estimating the posterior probability $z$ that a given observation $x$ belongs to a given class $k$. This is done by using Eq.~(\ref{eq12}). Third, the step {\bf 2)} can be regarded as an M-Step, i.e., maximizing the posterior probability to find the optimal mean $\mu_{k}$ or $\hat{\mu}_{k}$ and diagonal covariance $diag(\sigma_{k})$ or $diag(\hat{\sigma}_{k})$. This is done according to Eqs.~(\ref{eq13}) and (\ref{eq14}). Finally, the above two steps are repeated until convergence. Here, we denote the number of iteration as a hyper-parameter $n_{iter}$ and empirically find that setting it to 6 is sufficient to converge. For clarity, we summarize the improved EM-based method in Appendix~\ref{appendix_b}.

\subsubsection{Theoretic Analysis}
\label{section_3_5_5}
Here, we provide a brief theoretic analysis on the Gaussian-based prototype fusion strategy described in Section~\ref{section_3_5_2}. By the strategy, we can obatin five estimations, i.e., ${\hat \mu}'_k$, $\mu_k$, ${\hat \mu}_k$, $p_k$, and ${\hat p}_k$. Next, we analyze why the prototypes ${\hat \mu}'_k$ produced by the prototype fusion strategy are better.

\noindent {\bf Proposition 1.} $\mu_k$ (${\hat \mu}_k$) is more representative than $p_k$ (${\hat p}_k$).

\noindent \emph{Proof.}
We take $\mu_{k}$ and $p_k$ as an example to prove the Proposition 1. The proof for ${\hat \mu}_k$ and ${\hat p}_k$ is similar. Let us first revisit how are the variables $\mu_k$ and $\sigma_{k}$ estimated. In these EM-based fusion parameter estimation methods, the estimation of $\mu_k$ and $\sigma_{k}$ is regarded as a fitting problem of observation data $x \in \mathcal{S} \cup \mathcal{Q}$ with a $N$-components Gaussian mixture model. Thus, our goal is to optimize the $N$-components parameters $\psi^{t+1} = \{\mu_{k}^{t+1}, \sigma_{k}^{t+1}\}_{k=0}^{N-1}$ iteratively by maximizing the log-likelihood $L(\psi^{t})$:
\begin{small}
\begin{equation}
	\begin{aligned}
		\mathop{\max}\limits_{\psi^{t}} L(\psi^{t}) &= log\ \prod_{x \in \mathcal{S} \cup \mathcal{Q}} P(x|\psi^{t}) 
		\\&= \sum_{x \in \mathcal{S} \cup \mathcal{Q}} log\ \sum_{k=0}^{N-1} P(x, k|\psi^{t}),
	\end{aligned} 
	\label{eq16}
\end{equation}
\end{small}where $k$ is the label of $k$-th Gaussian components. As our solution follows the EM optimization, we have $L(\psi^{t+1}) \ge L(\psi^{t})$. This means that each iteration of the improved EM-based algorithm increases the log likelihood $L(\psi_{t})$, i.e., the parameters $\psi_{t+1}$ is more effective than $\psi_{t}$ for fitting observation data $x \in \mathcal{S} \cup \mathcal{Q}$. Thus, the variable $\mu_k$ obtained by the improved EM-based methods is more representative than the initial variable $p_k$.

\noindent {\bf Proposition 2.} ${\hat \mu}'_k$ is more representative than $\mu_k$, and ${\hat \mu}_k$.

\noindent \emph{Proof.} Let us revisit the fused prototype distribution, i.e., the posterior MGD $N({\hat \mu}'_k, diag(\hat{\sigma}_{k}'^2))$. Here, $\hat{\sigma}_{k}'^2=\frac{\sigma_{k}^2 \odot \hat{\sigma}_{k}^2}{\hat{\sigma}_{k}^2 + \sigma_{k}^2}$ denotes the estimation variance of prototypes ${\hat \mu}'_k$ (Note that we assume the covariance is diagonal). Then, we have the two inequalities since these terms $\frac{\sigma_{k}^4}{\hat{\sigma}_{k}^2 + \sigma_{k}^2}$ and $\frac{\hat{\sigma}_{k}^4}{\hat{\sigma}_{k}^2 + \sigma_{k}^2}$ are always greater than or equal to 0:
\begin{equation}
	\hat{\sigma}_{k}'^2 = \frac{\sigma_{k}^2 \odot \hat{\sigma}_{k}^2}{\hat{\sigma}_{k}^2 + \sigma_{k}^2} = \sigma_{k}^2 - \frac{\sigma_{k}^4}{\hat{\sigma}_{k}^2 + \sigma_{k}^2} \leq \sigma_{k}^2,
	\label{eq20}
\end{equation}
\begin{equation}
	\hat{\sigma}_{k}'^2 = \frac{\sigma_{k}^2 \odot \hat{\sigma}_{k}^2}{\hat{\sigma}_{k}^2 + \sigma_{k}^2} = \hat{\sigma}_{k}^2 - \frac{\hat{\sigma}_{k}^4}{\hat{\sigma}_{k}^2 + \sigma_{k}^2} \leq \hat{\sigma}_{k}^2,
	\label{eq21}
\end{equation}
where the right equation is satisfied only when $\sigma_{k}^2$ or $\hat{\sigma}_{k}^2$ is zero. The Eqs.~\ref{eq20} and \ref{eq21} imply that the variance of prototypes ${\hat \mu}'_k$ decreases for each class $k$ after fusing $\mu_k$ and ${\hat \mu}_k$. Thus, ${\hat \mu}'_k$ is more representative than $\mu_k$ and ${\hat \mu}_k$.

Based on the above propositions 1 and 2, we know that ${\hat \mu}'_k$ is more representative than $\mu_k$, ${\hat \mu}_k$, $p_k$, and ${\hat p}_k$. Hence, we take the mean $\hat{\mu}'_{k}$ as the final fused prototype $\hat{p}'_{k}$.

\section{Performance Evaluation}
\label{section_4}
In this section, we evaluate the proposed framework on general and fine-grained few-shot classification tasks, and then discuss the experiment results and present our statistical analysis, ablation study, and visualization in details. 

\subsection{Datasets and Settings}
\noindent {\bf MiniImagenet.} The data set is a subset of ImageNet, which includes 100 classes and each class consists of 600 images. Following \cite{XueW20}, we split the data set into 64 classes for training, 16 classes for validation, and 20 classes for test, respectively. The class parts/attributes are extracted from WordNet by using the relation of ``part\_holonyms()''. 

\noindent {\bf TieredImagenet.} 
The data set is another subset of ImageNet, which includes 608 classes and each class contains about 1200 images \cite{RenTRSSTLZ18}. It is first partitioned into 34 high-level classes, and then split into 20 classes for training, 6 classes for validation, and 8 classes for test, respectively. Similarly, the class parts/attributes are also extracted from WordNet by using the relation of ``part\_holonyms()''.

\noindent {\bf CUB-200-2011.}
The data set is a fine-grained classification data set, which includes 200 classes and contains about 11,788 images. Following \cite{zou2020compositional}, we split the data set into 100 classes for training, 50 classes for validation, and 50 classes for test, respectively. Different from miniImagenet and tieredImagenet, its class parts/attributes have been manually labeled and made publicly available. 



\subsection{Implementation Details}
\noindent{\bf Architecture.}
Following \cite{chen2020new}, we employ ResNet12 as the feature extractor. In PATNet, we use a single-layer MLP with 512 units for the embedding layer, and a two-layer MLP with 512-dimensional hidden units for the mean module and diagonal covariance module, respectively. In ProtoComNet, we use a single-layer MLP with 256 units for the encoder, a two-layer MLP with a 300-dimensional hidden layer for the aggregator, and a two-layer MLP with 512-dimensional hidden layers for the decoder. Here, ReLU is used as the activation function for all network. The number of iteration, namely $n_{iter}$, is set to 6 for GaussFusion. 

\noindent {\bf Training Details.}
We first pre-train the feature extractor with 100 epochs on base classes via an SGD with momentum of 0.9 and weight decay of 0.0005. The learning rate is initially set to 0.1, and then decayed by 0.1 at epochs 60, 80, and 90, respectively. Second, we train the PATNet with 20000 epochs by using an Adam with weight decay of 0.0005. The learning rate is initially set to 0.001, and then decayed by 0.1 at 10000 epochs. Third, we train the ProtoComNet with 100 epochs in an episodic manner by using an SGD with momentum of 0.9 and weight decay of 0.0005. The learning rate is initially set to 0.1, and then changed at epochs 15, 40, and 80. Finally, we fine-tune all modules with 40 epochs in an episodic manner. The learning rate is initially set to 0.01, and then decayed by 0.1 at epochs 15, 25, and 30. 

\noindent{\bf Evaluation.} We conduct few-shot classification on 600 randomly sampled episodes from the test set and report the mean accuracy together with the 95\% confidence interval. In each episode, we randomly sample 15 query images per class for evaluation in 5-way 1-shot/5-shot tasks. 

\subsection{Discussion of Results}
For a comparison, some state-of-the-art approaches are also applied to the few-shot classification and few-shot fine-grained classification tasks as baselines. These methods can be roughly from six types, \emph{i.e.}, metric-based, optimization-based, semantics-based, attribute-based, graph-based, and pre-training based approaches. For a fair comparison, we employ the MeanFusion and GaussFusion strategy to evaluate the performance of our framework on inductive and transductive FSL seting, respectively.

\begin{table*}[t]
	\caption{Experiment results on the miniImagenet and tieredImagenet data sets. The best results are highlighted in bold. In. and Tran. indicate inductive and transductive FSL setting, respectively. `\_' denotes the absent results in  original paper. }\smallskip
	\centering
	\smallskip\scalebox
	{0.95}{\begin{tabular}{l|l|c|c|c|c|c|c}
			\hline
			\multicolumn{1}{l|}{\multirow{2}{*}{Setting}}&
			\multicolumn{1}{l|}{\multirow{2}{*}{Method}}&
			\multicolumn{1}{c|}{\multirow{2}{*}{Type}}& \multicolumn{1}{c|}{\multirow{2}{*}{Backbone}}&
			\multicolumn{2}{|c|}{miniImagenet} & \multicolumn{2}{|c}{tieredImagenet} \\ 
			\cline{5-8}
			& & & & 5-way 1-shot & 5-way 5-shot & 5-way 1-shot & 5-way 5-shot \\
			\hline \hline
			\multicolumn{1}{l|}{\multirow{21}{*}{In.}} 
			& RestoreNet \cite{XueW20} & Metric & ResNet18 &  $59.28 \pm 0.20\%$  & $- \pm -\%$ & $- \pm -\%$  & $- \pm -\%$ \\ 
			& ConstellationNet \cite{XuXWT21} & Metric & ResNet12 &  $64.89 \pm 0.23\%$  & $79.95 \pm 0.17\%$ & $- \pm -\%$  & $- \pm -\%$ \\ 
			& RAP-ProtoNet \cite{Hong_2021_CVPR} & Metric & ResNet10 &  $53.64 \pm 0.60\%$  & $74.54 \pm 0.45\%$ & $- \pm -\%$  & $- \pm -\%$ \\ 
			& MAML \cite{finn2017model} & Optimization & ResNet12 & $58.37 \pm 0.49\%$  & $69.76 \pm 0.46\%$ & $58.58 \pm 0.49\%$  & $71.24 \pm 0.43\%$ \\
			& MetaOptNet\cite{lee2019meta} & Optimization & ResNet12 &  $62.64 \pm 0.61\%$  & $78.63 \pm 0.46\%$ & $65.99 \pm 0.72\%$  & $81.56 \pm 0.53\%$ \\
			& ALFA \cite{BaikCCKL20} & Optimization & ResNet12 & $59.74 \pm 0.49\%$  & $77.96 \pm 0.41\%$ & $64.62 \pm 0.49\%$  & $82.48 \pm 0.38\%$ \\
			& AM3-TRAML \cite{boostingfew} & Semantics & ResNet12 &  67.10 $\pm$ 0.52 $\%$  & 79.54 $\pm$ 0.60\% & $- \pm -\%$  & $- \pm -\%$ \\
			& MultiSem \cite{babysteps2019} & Semantics & Dense-121 &  $67.3\%$  & \textbf{82.1}$\%$ & $- \pm -\%$  & $- \pm -\%$ \\
			& FSLKT \cite{zhimao2019few} & Semantics & ConvNet128 & $64.42 \pm 0.72\%$  & $74.16 \pm 0.56\%$ & $- \pm -\%$  & $- \pm -\%$ \\
			& CPDE \cite{zou2020compositional} & Attribute & ResNet12 &  63.21 $\pm$ 0.78\%  & 79.68 $\pm$ 0.82$\%$ & $- \pm -\%$  & $- \pm -\%$ \\
			& CFA \cite{Hu2019Weaklys} & Attribute & ResNet18 &  $58.50 \pm  0.80\%$  & $76.60 \pm 0.60\%$ & $- \pm -\%$  & $- \pm -\%$ \\
			& MetaBaseline \cite{chen2020new} & Pre-training & ResNet12 &  $63.17 \pm 0.23\%$  & 79.26 $\pm$ 0.17\% & $68.62 \pm 0.27\%$  & $83.29 \pm 0.18\%$ \\
			& Neg-Cosine \cite{LiuCLL0LH20} & Pre-training & ResNet12 &  $63.85 \pm 0.81\%$  & $81.57 \pm 0.56\%$ & $- \pm -\%$  & $- \pm -\%$ \\
			& CentAlign \cite{AfrasiyabiLG20} & Pre-training & ResNet18 &  $59.88 \pm 0.67\%$  & $80.35 \pm 0.73\%$ & $69.29 \pm 0.56\%$  & $85.97 \pm 0.49\%$ \\ 
			& DC \cite{yangfree2021} & Pre-training & WRN-28-10 &  $66.91 \pm 0.17\%$  & $80.74 \pm 0.48\%$ & \textbf{75.92} $\pm$ \textbf{0.60}$\%$  & \textbf{87.84} $\pm$ \textbf{0.65}$\%$ \\
			
			\cline{2-8}
			& Our Method (MeanFusion) & Pre-training & ResNet12 & \textbf{69.68} $\pm$ \textbf{0.76}$\%$ & \textbf{81.65} $\pm$ \textbf{0.54}$\%$ & 74.19 $\pm$ 0.90$\%$ & 86.09 $\pm$ 0.60$\%$ \\
			\hline
			\multicolumn{1}{l|}{\multirow{14}{*}{Trans.}} & SRestoreNet \cite{XueW20} & Metric & ResNet18 &  $61.14 \pm 0.22\%$  & $- \pm -\%$ & $- \pm -\%$  & $- \pm -\%$ \\
			& DPGN\cite{YangLZZZL20} & Graph & ResNet12 &  $67.77 \pm 0.32\%$  & 84.60 $\pm$ 0.43\% & $72.45 \pm 0.51\%$  & $87.24 \pm 0.39\%$ \\
			& EPNet\cite{RodriguezLDL20} & Graph & ResNet12 & $66.50 \pm 0.89\%$  & $81.06 \pm 0.60\%$ & $76.53 \pm 0.87\%$  & $87.32 \pm 0.64\%$ \\
			& MCGN\cite{Tang_2021_CVPR} & Graph & ConvNet256 & $67.32 \pm 0.43\%$  & $83.03 \pm 0.54\%$ & $71.21 \pm 0.85\%$  & $85.98 \pm 0.98\%$ \\
			& TIM-GD\cite{Boudiaf20} & Pre-training & ResNet18 & $73.9 \pm -\%$  & $\textbf{85.0} \pm -\%$ & $79.9 \pm -\%$  & $88.5 \pm -\%$ \\
			& TFT\cite{DhillonCRS20} & Pre-training & WRN-28-10 &  $65.73 \pm 0.68\%$  & $78.40 \pm 0.52\%$ & $73.34 \pm 0.71\%$  & $85.50 \pm 0.50\%$ \\ 
			& SIB\cite{HuMXSOLD20} & Pre-training & WRN-28-10 & $70.0 \pm 0.6\%$  & $79.2 \pm 0.4\%$ & $- \pm -\%$  & $- \pm -\%$ \\
			& LaplacianShot\cite{ZikoDGA20} & Pre-training & ResNet18 & $72.11 \pm 0.19\%$  & $82.31 \pm 0.14\%$ & $78.98 \pm 0.21\%$  & $86.39 \pm 0.16\%$ \\
			& RAP-LaplacianShot\cite{Hong_2021_CVPR} & Pre-training & ResNet12 & $74.29 \pm 0.20\%$  & $84.51 \pm 0.13\%$ & $- \pm -\%$  & $- \pm -\%$ \\
			& ICI\cite{WangXLZF20} & Pre-training & ResNet12 & $65.77 \pm -\%$  & $78.94 \pm -\%$ & $80.56 \pm -\%$  & $87.93 \pm -\%$ \\			 
			& BD-CSPN \cite{YaohuiWang_pr} & Pre-training & ResNet12 &  65.94$\%$  & 79.23$\%$ & 76.17$\%$  & 85.70$\%$ \\
			\cline{2-8}
			& Conference Version \cite{zhang2021prototype} & Pre-training & ResNet12 & 73.13 $\pm$ 0.85$\%$ & 82.06 $\pm$ 0.54$\%$ & 81.04 $\pm$ 0.89$\%$ & 87.42 $\pm$ 0.57$\%$ \\
			& Our Method (EM) & Pre-training & ResNet12 & 75.35$\pm$ 0.87$\%$ & 83.46 $\pm$ 0.58$\%$ & 81.40 $\pm$ 0.96$\%$ & 88.15 $\pm$ 0.59$\%$ \\
			& Our Method (Improved EM) & Pre-training & ResNet12 & \textbf{79.01} $\pm$ \textbf{0.89}$\%$ & \textbf{84.18} $\pm$ \textbf{0.56}$\%$ & \textbf{83.06} $\pm$ \textbf{1.00}$\%$ & \textbf{88.60} $\pm$ \textbf{0.57}$\%$ \\
			\hline
	\end{tabular}}
	\label{table1}
\end{table*}

\emph{1)} {\bf In few-shot classification.} Table~\ref{table1} shows the results of our method and the baseline methods on miniImagenet and tieredImagenet. It can be found that our method achieves superior performance on  both inductive and transductive FSL settings. Specifically, in inductive FSL, compared with the metric-based approaches, our method better exploits the power of pre-training by learning to complete prototypes. The results show our method is more effective, with an improvement of 4\% $\sim$ 10\%. It is worth noting that our method also beats RestoreNet, which also adopts the strategy of prototype learning. This demonstrates the proposed prototype completion is more effective. Compared with the optimization-based methods (\emph{e.g.,} ALFA), our method achieves 3\% $\sim$ 9\% higher accuracy. Different from these methods, we focus on metric-based FSL framework, but targets at learning representative prototypes. As for the semantics and attribute-based approaches, they also leverage the external knowledge. However, our method utilizes the knowledge to learn to complete prototypes, instead of to combine modality or to learn the feature extractor. The result validates the superiority of our manner to incorporate the external knowledge. Note that our method achieves competitive performance with the MultiSem method on 5-shot tasks on miniImagenet. We would like to emphasize that this is because MultiSem leverages a more complex backbone, namely the Dense-121 with 121 layers, instead of ResetNet12 in our model.

Finally, from the results of the pre-training based apporaches, we have the following observations. (\romannumeral1) Our method exceeds the MetaBaseline method by a large margin, around 3\%$\sim$7\% (1-shot) and 2\% $\sim$ 4\% (5-shot). This verifies our motivation that estimating more accurate prototypes is more effective than fine-tuning feature extractor during meta-learning. Besides, the improvement of performance on 1-shot tasks is more obvious than on 5-shot tasks. This is reasonable because the problem of inaccurate estimation of prototypes on 1-shot is more remarkable than 5-shot tasks. (\romannumeral2) Our method outperforms Neg-Cosine and CentAlign, by around 1\% $\sim$ 5\%. This is because our method focuses on estimating more representative prototypes, instead of pre-training strategy or generating more training samples. (\romannumeral3) Our method exceeds DC method by around 1\% $\sim$ 3\% on miniImagenet, while performs slightly worse than DC on tieredImagenet. The reason is that the DC method leverags a deeper backbone WRN-28-10 instead of ResNet12 and a complex power transformations for image representation. 

\begin{table}[t]
	\caption{Experiment results on the CUB-200-2011 data set. 
		The best results are highlighted in bold. In. and Tran. indicate inductive and transductive FSL setting, respectively. 
	}\smallskip
	\centering
	\smallskip\scalebox
	{0.85}{\begin{tabular}{l|l|c|c}
			\hline
			\multicolumn{1}{l|}{\multirow{2}{*}{Setting}} & \multicolumn{1}{l|}{\multirow{2}{*}{Method}} & \multicolumn{2}{|c}{CUB-200-2011}\\ 
			\cline{3-4}
			& & 5-way 1-shot & 5-way 5-shot\\
			\hline \hline
			\multicolumn{1}{l|}{\multirow{11}{*}{In.}} & RestoreNet \cite{XueW20} & $74.32 \pm 0.91\%$  & $- \pm -\%$ \\
			& RAP-ProtoNet \cite{Hong_2021_CVPR} & $75.17 \pm 0.63\%$  & $88.29 \pm 0.34\%$ \\
			& MAML \cite{finn2017model} & $55.92 \pm 0.95\%$  & $72.09 \pm 0.76\%$ \\
			& MultiSem \cite{babysteps2019} & $76.1\%$  & $82.9\%$ \\ 
			& CPDE \cite{zou2020compositional} & 80.11 $\pm$ 0.34 $\%$  & 89.28 $\pm$ 0.33$\%$ \\
			& CFA \cite{Hu2019Weaklys} & $73.90 \pm 0.80\%$  & $86.80 \pm 0.50\%$ \\
			& Neg-Cosine \cite{LiuCLL0LH20} &  $72.66 \pm 0.85\%$  & $89.40 \pm 0.43\%$ \\
			& CentAlign \cite{AfrasiyabiLG20} &  $74.22 \pm 1.09\%$  & $88.65 \pm 0.55\%$ \\
			& DC \cite{Hu2019Weaklys} & $77.22 \pm 0.14\%$  & $89.58 \pm 0.27\%$ \\
			\cline{2-4}
			& Our Method & \textbf{88.99} $\pm$ \textbf{0.58}$\%$ & \textbf{94.05} $\pm$ \textbf{0.34}$\%$ \\
			\hline
			\multicolumn{1}{l|}{\multirow{10}{*}{Trans.}} & SRestoreNet \cite{XueW20} & $76.85 \pm 0.95\%$  & $- \pm -\%$ \\
			& EPNet \cite{RodriguezLDL20} &$82.85 \pm 0.81\%$  &$91.32 \pm 0.41\%$\\
			& ICI \cite{WangXLZF20} & $87.87\%$  & $92.38\%$ \\
			& TIM-GD \cite{Boudiaf20} & 82.2 $\%$  & 90.8 $\%$ \\
			& LaplacianShot \cite{ZikoDGA20} & $80.96\%$  & $88.68\%$ \\
			& RAP-LaplacianShot \cite{Hong_2021_CVPR} & 83.59 $\pm$ 0.18$\%$  & 90.77 $\pm$ 0.10$\%$ \\
			& BD-CSPN \cite{YaohuiWang_pr} & 84.90 $\%$  & 90.22$\%$ \\
			\cline{2-4}
			& Conference Version \cite{zhang2021prototype} & 93.20 $\pm$ 0.45$\%$ & 94.90 $\pm$ 0.31$\%$ \\
			& Our Method (EM) & 92.35 $\pm$ 0.62$\%$ & 95.08 $\pm$ 0.31$\%$ \\
			& Our Method (Improved EM) & \textbf{93.78} $\pm$ \textbf{0.55}$\%$ & \textbf{95.19} $\pm$ \textbf{0.30}$\%$ \\
			\hline
	\end{tabular}}
	\label{table2}
\end{table}

In transductive FSL setting, SRestoreNet is very related with our method, which also explores the query samples to restore prototypes. However, different from it, we leverage the query samples to estimate the prototype distribution and then to fuse prototypes. The result validates the superiority of our method. Compared with the graph-based approaches, our method obtains competitive classification performance, especially in 1-shot tasks. This is because our method exploits unlabeled data to combine mean-based and completed prototypes, instead of propagate embedding or labels. Finally, from the results of the pre-training based apporaches, we have the following observations. (\romannumeral1) Compared with the best results of pre-training based methods (TIM-GD, SIB, LaplacianShot, and ICI), our method obtains 1\% $\sim$ 6\% higher accuracy, which further validates the superiority of learning representative prototypes. (\romannumeral2) Our method outperforms BD-CSPN, by around 5\% $\sim$ 14\%. The DB-SCPN method also leverages unlabeled samples, but they only focus on pre-training and ignore the advantange of meta-learning. Different from it, we introduce a meta-learner, learning to complete prototypes, to explore the power of pre-training further. Besides, the improvement of performance on 1-shot tasks is more obvious than on 5-shot tasks. This is reasonable because the problem of inaccurate estimation of prototypes on 1-shot is more remarkable than 5-shot tasks. (\romannumeral3) Compared with the conference version \cite{zhang2021prototype}, the extended version (EM-based and Improved EM-based) exceeds it by 1\% $\sim$ 6\%. The main reason is that we explore unseen parts/attributes and enhance the GaussFusion by introducing an iterable parameter estimation algorithm. (\romannumeral4) our improved EM-based method perform best in all extended methods, thus it is used in subsequent discussion.

\begin{figure}[!t]
	\centering
	\includegraphics[width=0.85\columnwidth]{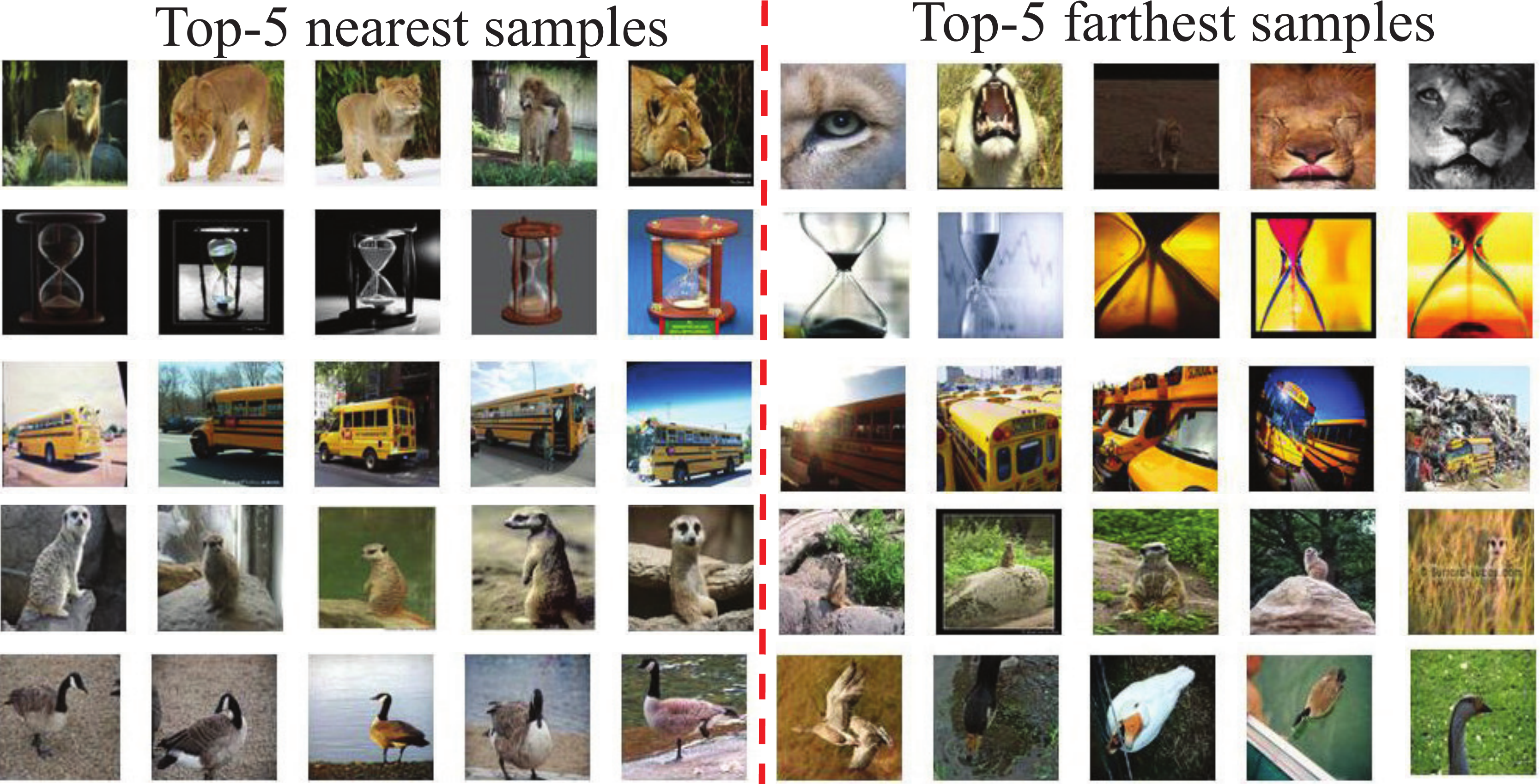}  
	\caption{Top-5 nearest and farthest samples from centers.}
	\label{fig4}
\end{figure}

\emph{2)} {\bf In few-shot fine-grained classification.} Table~\ref{table2} summarizes the results on CUB-200-2011, which lead to similar observations as those in Table~\ref{table1}. We observe that our method (\romannumeral1) also achieves superior performance over state-of-the-art methods with an improvement of 4\% $\sim$ 5\% (inductive FSL) and 4\% $\sim$ 6\% (transductive FSL); (\romannumeral2) exceeds the conference version around 1\%; (\romannumeral3) obtains almost consistent performance on 1-shot and 5-shot tasks, while the improvements on 1-shot task over baselines are more significant than on 5-shot. The results on few-shot fine-grained classification tasks further verify the effectiveness of the proposed method, especially for 1-shot classification tasks.

\subsection{Statistical Analysis}
In this subsection, we conduct additional statistical experiments to answer the following four questions:

{\emph{1)} {\bf Is our idea reasonable on realistic data?} We randomly select five classes from the novel classes of miniImageNet and retrieve top-5 nearest and farthest samples from its ground-truth class center in the feature space. As shown in Fig.~\ref{fig4}, the nearest images are more complete; however, the farthest samples are missing partial parts/attributes due to its incompleteness, noise background, or obscured details.

\begin{table}
	\caption{The cosine similarity between the estimated and real prototypes
		on 1000 episodes (5-way 1-shot) of miniImagenet, tieredImagenet, and CUB-200-2011. 
		$d(x, y)$ denotes the cosine simiarity of vectors $x$ and $y$.}\smallskip
	\centering
	\smallskip\scalebox
	{0.95}{\begin{tabular}{l|c|c|c}
			\hline
			Methods & $d(p_k, p^{real}_k)$ & $d(\hat{p}_k, p^{real}_k)$ & $d(\hat{p}'_k, p^{real}_k)$\\
			\hline \hline
			\multicolumn{4}{c}{\multirow{1}{*}{miniImagenet}} \\
			\hline
			SRestoreNet  & 0.55 & 0.78 & 0.79\\
			BD-CSPN  & 0.55 & - & 0.67\\
			Conference Version \cite{zhang2021prototype} & 0.55 & 0.71 & 0.90\\
			Our Method & 0.55 & 0.77 & 0.96\\
			\hline \hline
			\multicolumn{4}{c}{\multirow{1}{*}{tieredImagenet}} \\
			\hline
			SRestoreNet  & 0.72 & 0.86 & 0.91\\
			BD-CSPN  & 0.72 & - & 0.83\\
			Conference Version \cite{zhang2021prototype} & 0.72 & 0.84 & 0.95\\
			Our Method & 0.72 & 0.85 & 0.97\\
			\hline \hline
			\multicolumn{4}{c}{\multirow{1}{*}{CUB-200-2011}} \\
			\hline
			SRestoreNet  & 0.68 & 0.83 & 0.89\\
			BD-CSPN  & 0.68 & - & 0.79\\
			Conference Version \cite{zhang2021prototype} & 0.68 & 0.77 & 0.95\\
			Our Method & 0.68 & 0.80 & 0.98\\
			\hline
	\end{tabular}}
	\label{table3}
\end{table} 

\begin{figure}[!t]
	\centering
	\subfigure[miniImagenet]{ 
		\label{fig2_a} 
		\includegraphics[width=0.80\columnwidth]{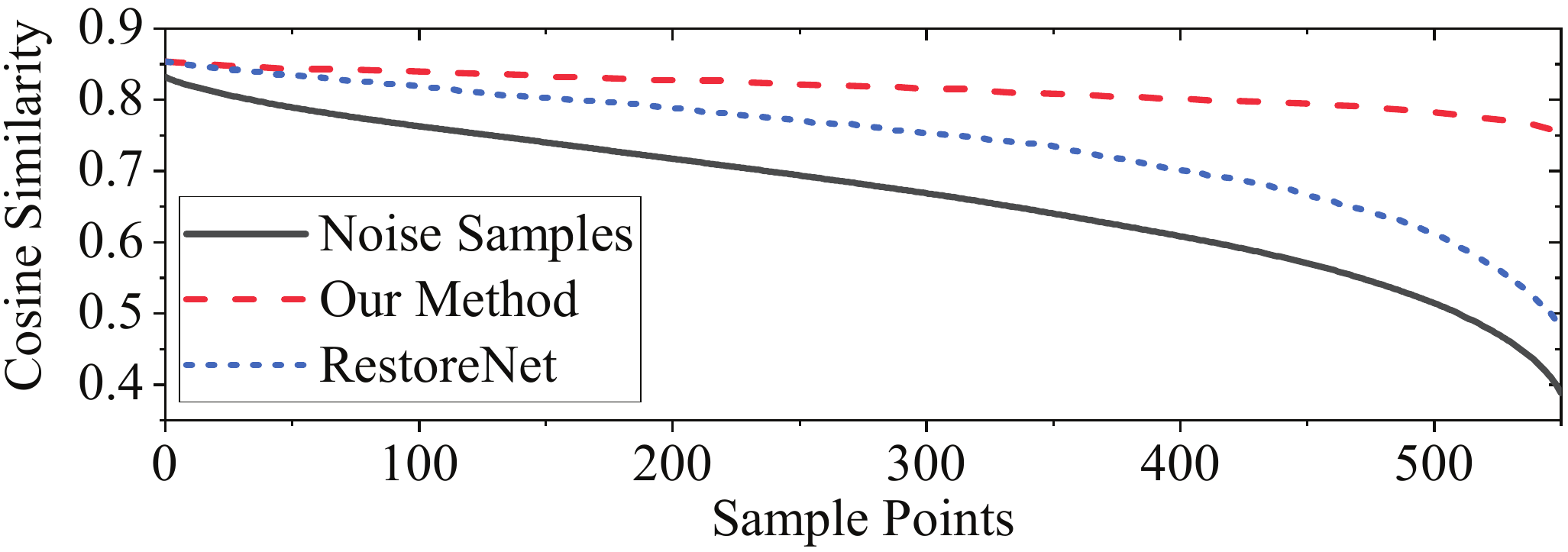}}
	\subfigure[tieredImagenet]{ 
		\label{fig2_b} 
		\includegraphics[width=0.85\columnwidth]{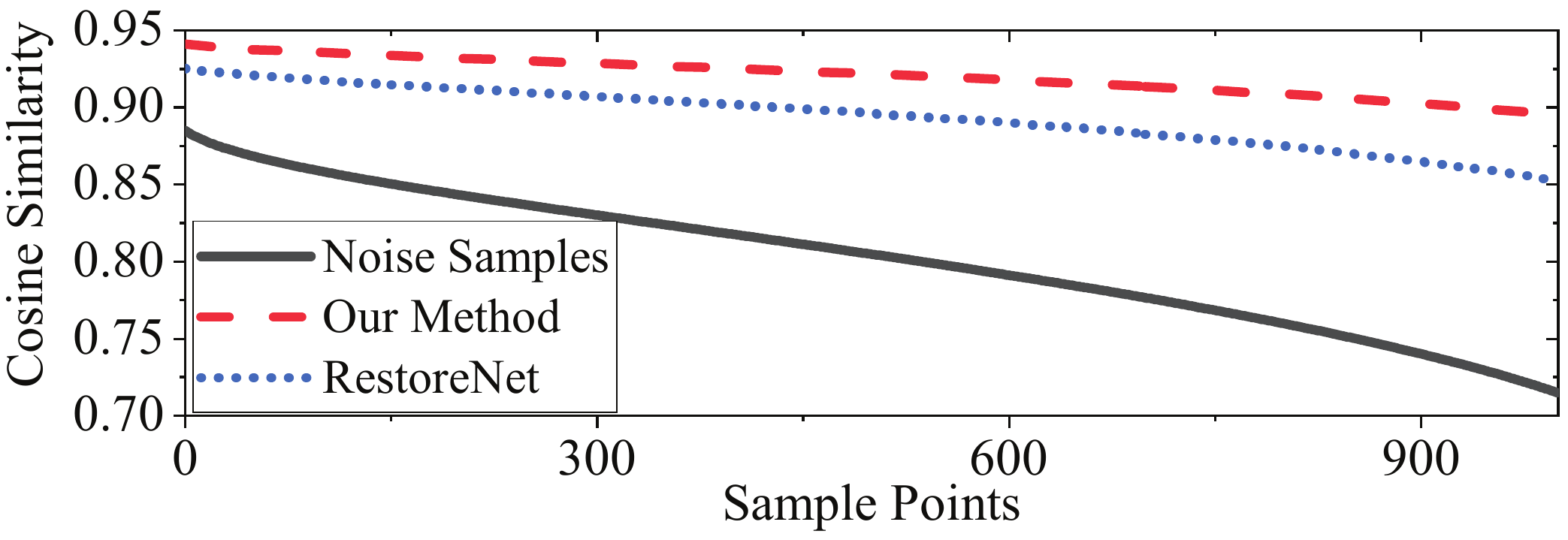}}
	\subfigure[CUB-200-2011]{ 
		\label{fig2_c} 
		\includegraphics[width=0.85\columnwidth]{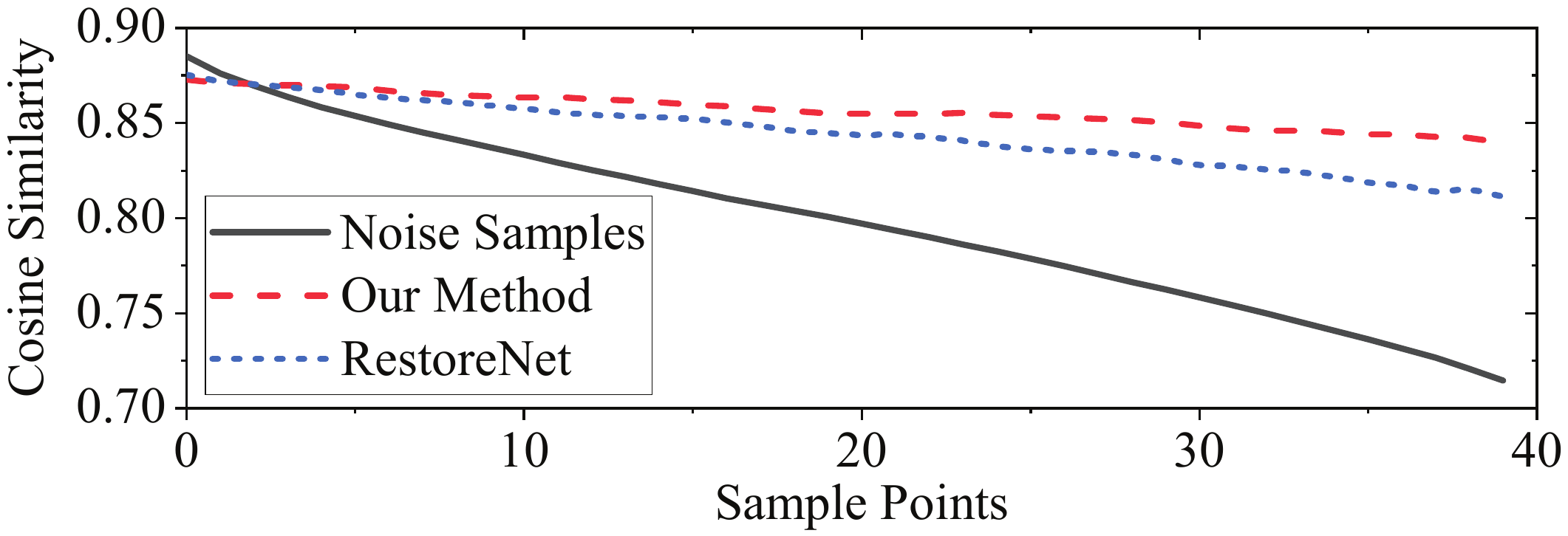}}
	\caption{Performance analysis of ProtoComNet on three datasets.}
	\label{fig2_0}
\end{figure}

{\emph{2)} {\bf Does our method obtain more accurate prototypes?} We calculate the average cosine similarity between the estimated prototypes and the real prototypes on 1000 episodes (5-way 1-shot) on miniImagenet, tieredImagenet, and CUB-200-2011. Three results including the mean-based ($p_k$), the restored/completed ($\hat{p}_k$) and the fused prototypes ($\hat{p}'_k$) are reported. For a fair comparison, we report the results of SRestoreNet, FSLKT, and BD-CSPN as the baselines. As shown in Table~\ref{table3}, the results show that our method obtains more accurate prototypes than these baselines and the conference version \cite{zhang2021prototype}. Note that the prototypes $\hat{p}_k$ from SRestoreNet is better than our method. This is reasonable because they leverage unlabeled samples before restoring prototypes. However, we exploit them after completing prototypes.

{\emph{3)} {\bf Is our method effective for the samples far away from its class center?} On the novel classes of miniImageNet, tieredImagenet, and CUB-200-2011, we calculate the cosine similarity between each noise image and its class center and sort them in descending order (\emph{i.e.}, the larger the sample number is, the farther away it is from the class center). Then, we take the noise images as inputs to predict the prototypes by using our method and RestoreNet, respectively. The cosine similarity between predicted prototypes and real class centers is shown in Fig.~\ref{fig2_0}. Note that (\romannumeral1) we smoothen the curve through moving average with 50 samples; (\romannumeral2) we show the average results for all novel classes. From the results of the above three datasets, we observe our method achieves more accurate prototypes than RestoreNet and the improvement becomes larger as the samples are farther away from its center. This means that our method can recover representative prototypes, especially when they are far away from their ground-truth centers.

\begin{figure}[!t]
	
	\centering
	\subfigure[miniImagenet 5-way 1-shot]{ 
		\label{fig6a} 
		\includegraphics[width=0.43\columnwidth]{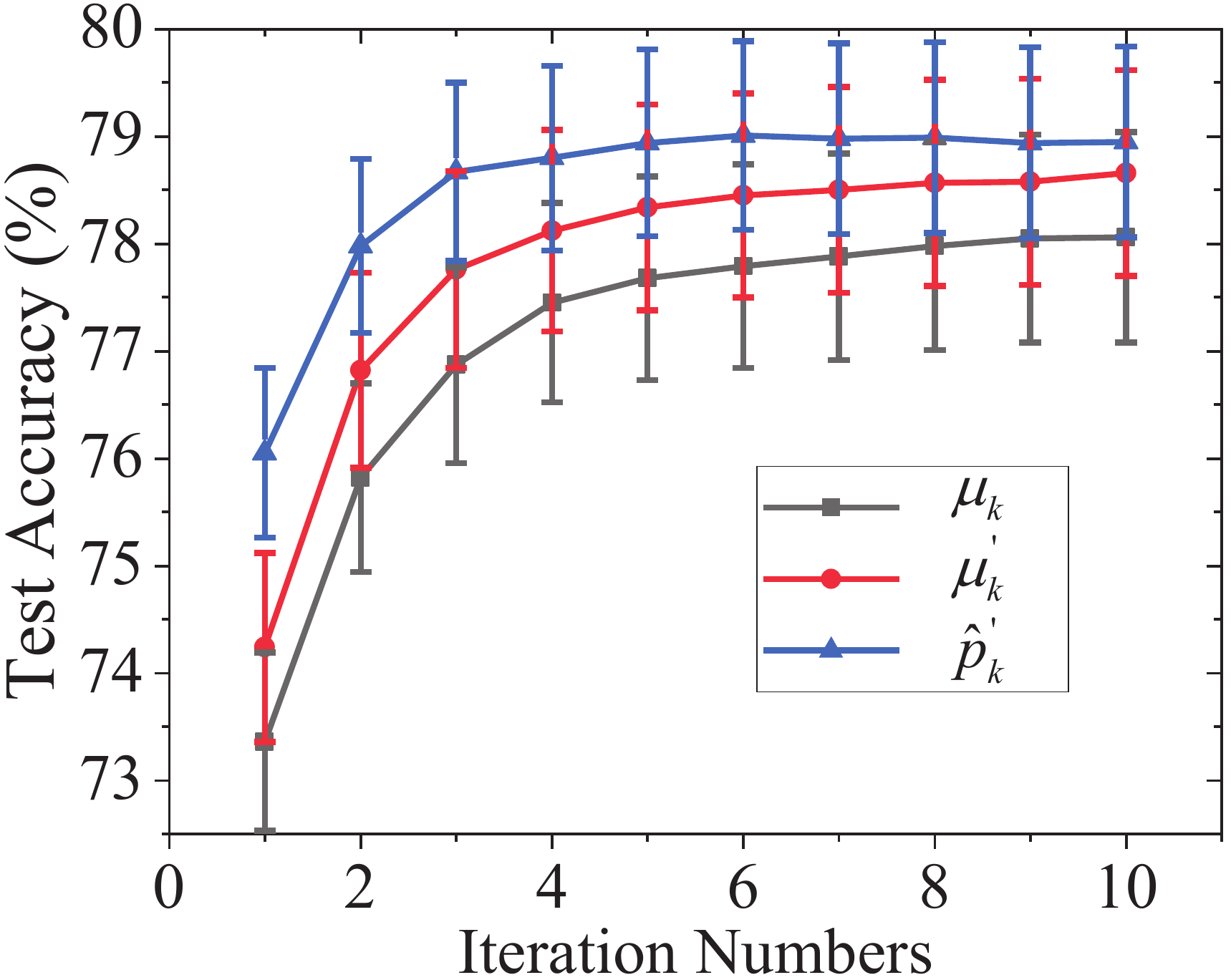}}
	\subfigure[miniImagenet 5-way 5-shot]{ 
		\label{fig6b} 
		\includegraphics[width=0.43\columnwidth]{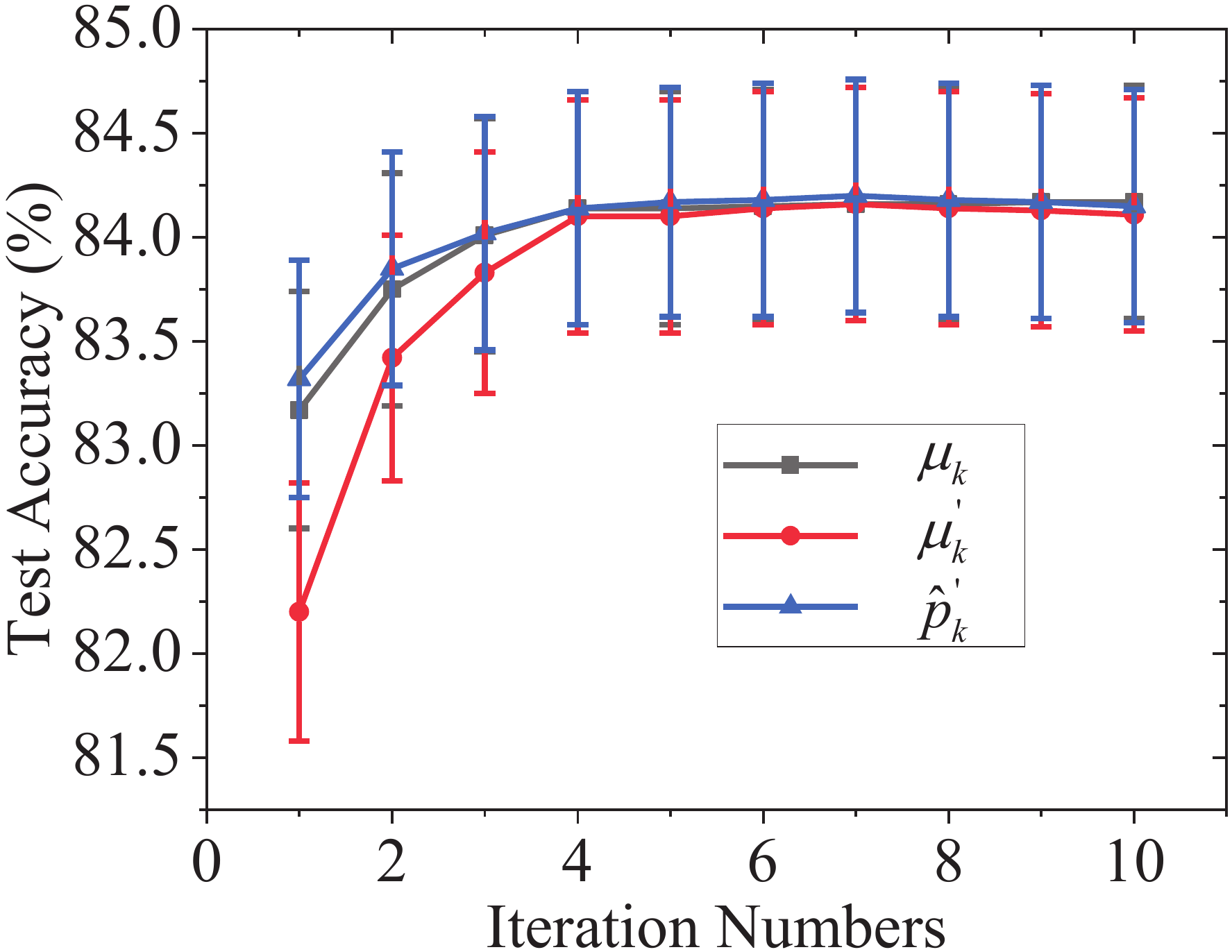}}
	\subfigure[tieredImagenet 5-way 1-shot]{ 
		\label{fig6c} 
		\includegraphics[width=0.44\columnwidth]{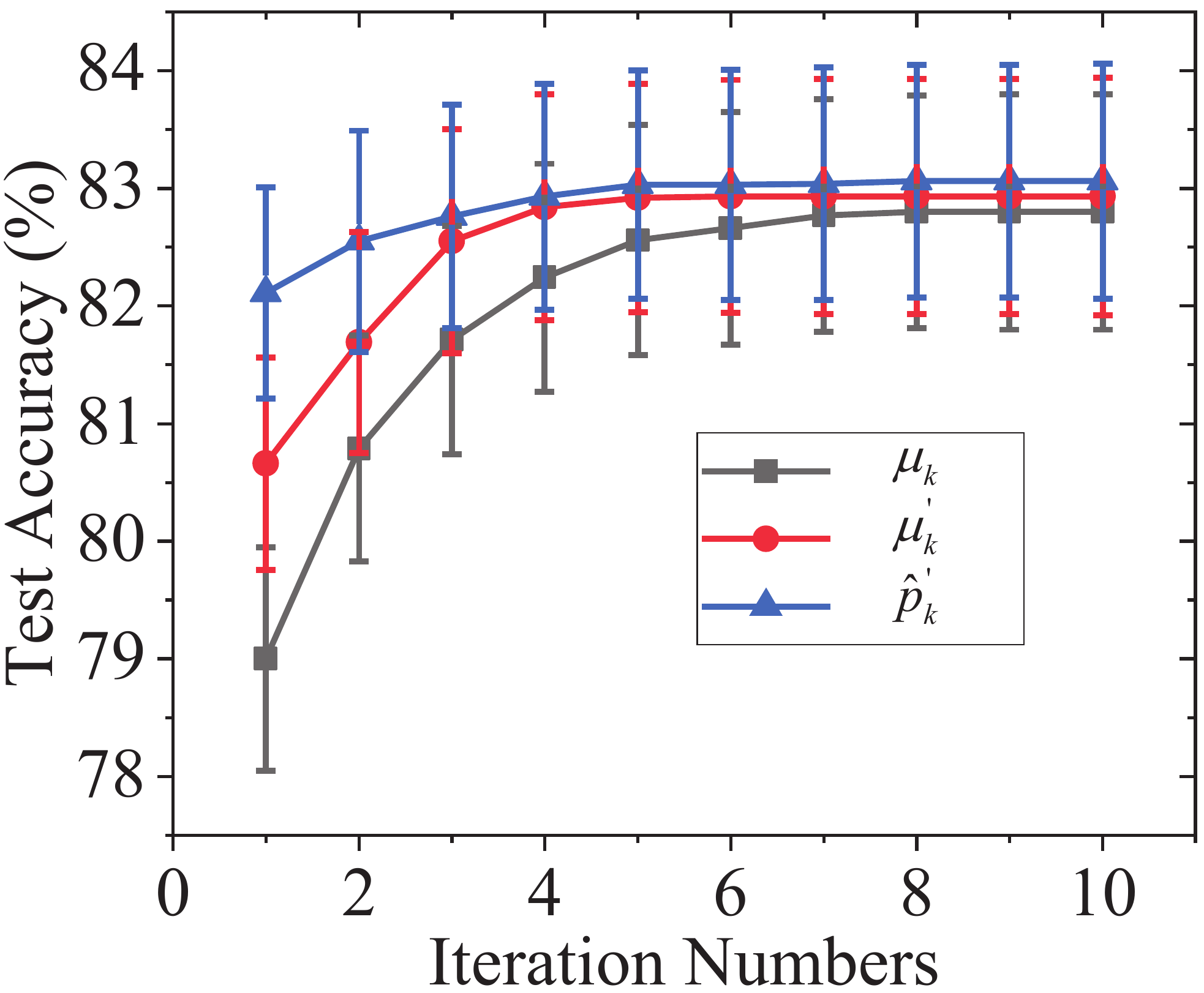}}
	\subfigure[tieredImagenet 5-way 5-shot]{ 
		\label{fig6d} 
		\includegraphics[width=0.44\columnwidth]{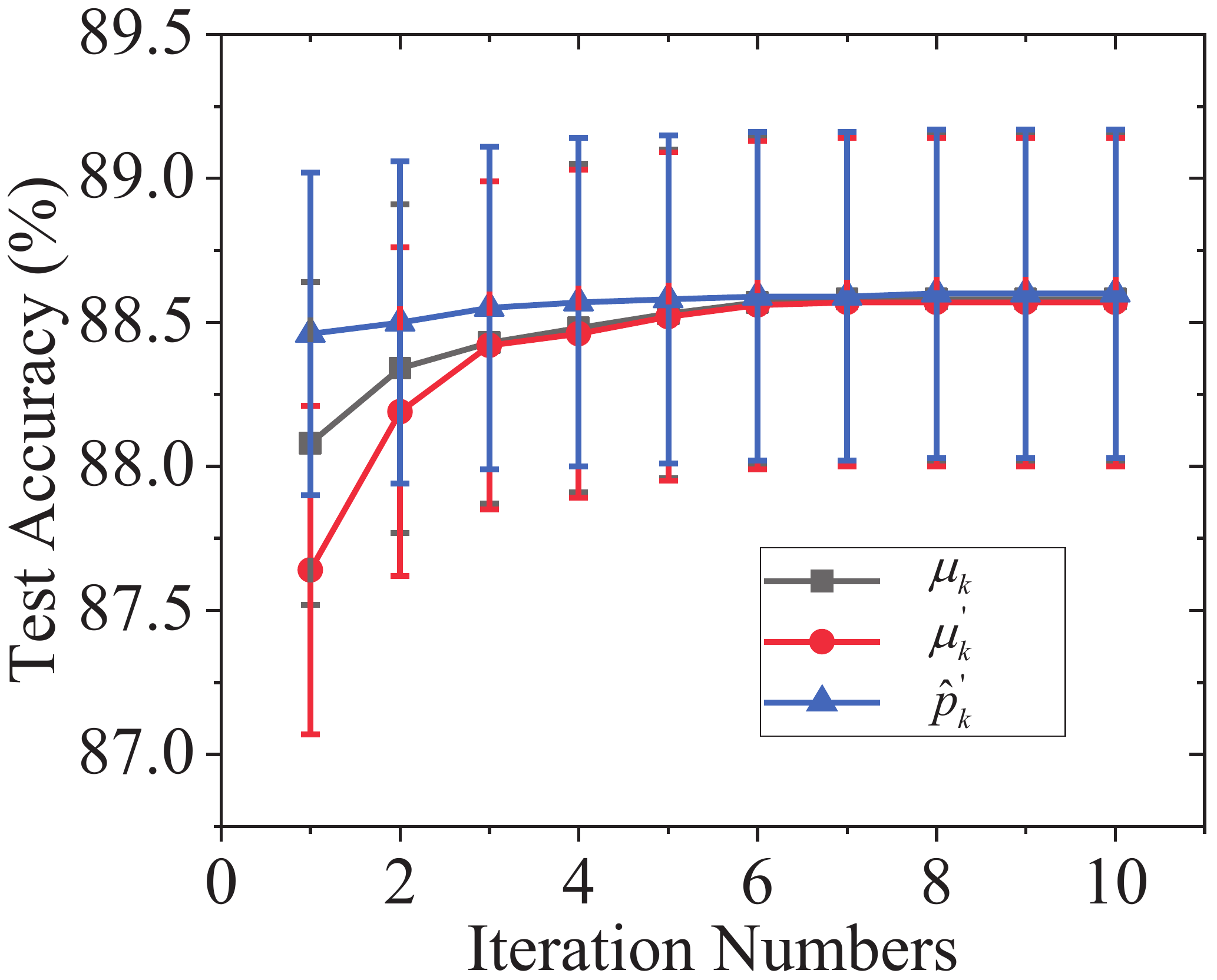}}
	\subfigure[CUB-200-2011 5-way 1-shot]{ 
		\label{fig6e} 
		\includegraphics[width=0.45\columnwidth]{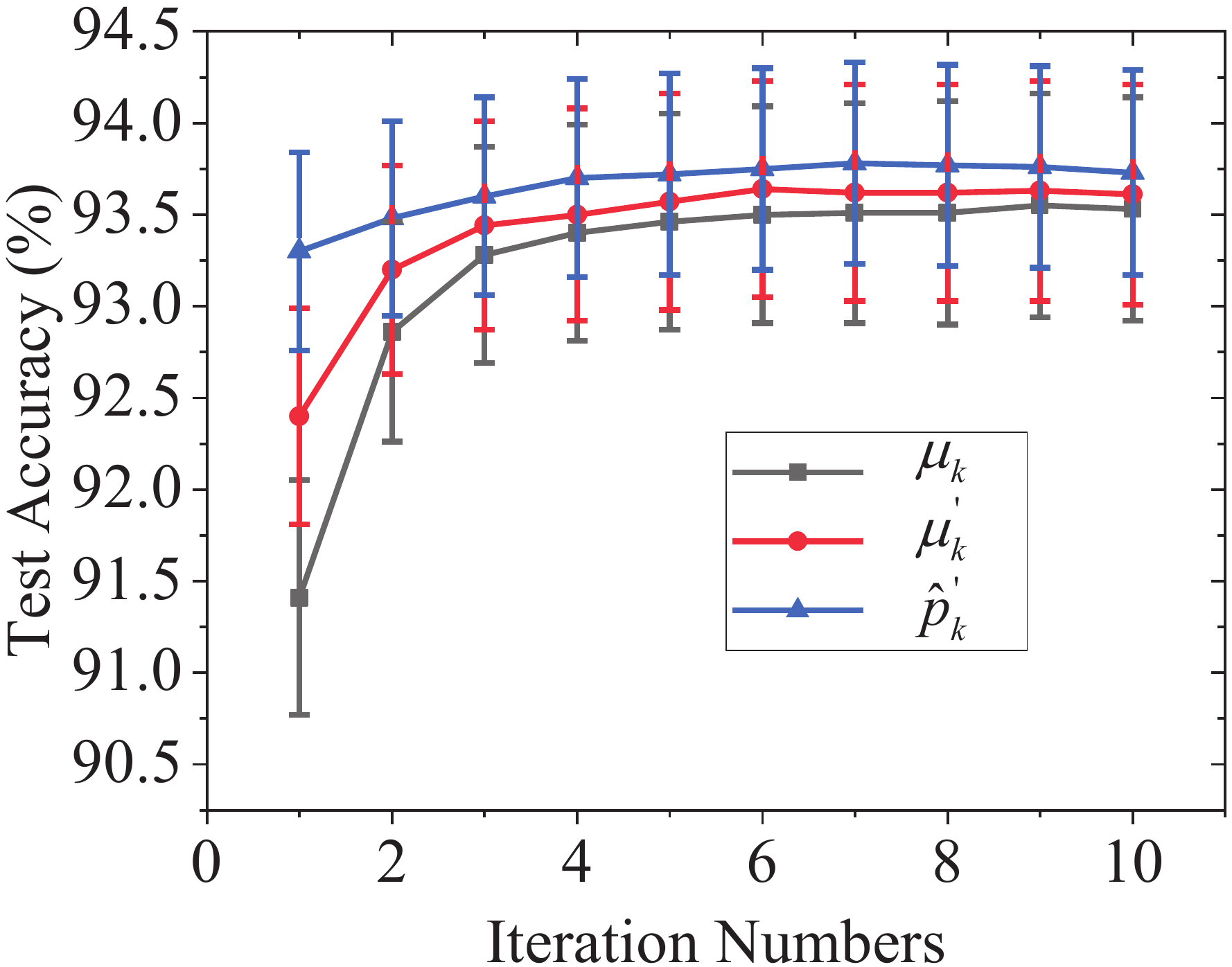}}
	\subfigure[CUB-200-2011 5-way 5-shot]{ 
		\label{fig6f} 
		\includegraphics[width=0.45\columnwidth]{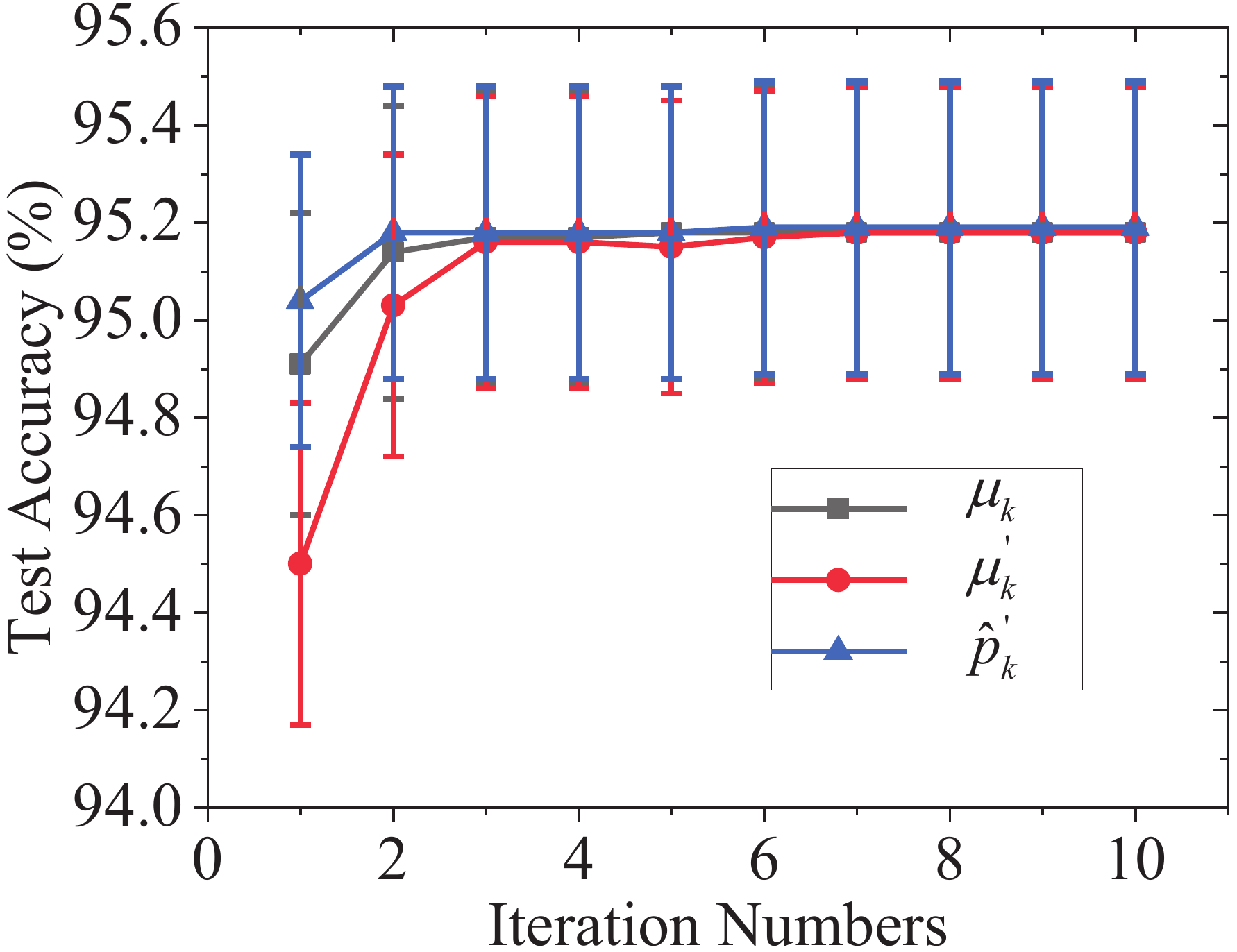}}
	\caption{Performance of GaussFusion with different iterations on 5-way 1/5-shot tasks of miniImagenet, tieredImagenet and CUB-200-2011.}
	
	\label{fig6}
\end{figure} 

{\emph{4)} {\bf How set the number of iterations $n_{iter}$ for GaussFusion with improved EM-based method?} To find the optimal $n_{iter}$, we conduct experiments on 5-way 1-shot and 5-shot tasks of miniImagenet, tieredImagenet, and CUB-200-2011, respectively, and report the test accuracy of the proposed method with different $n_{iter}$. The results are shown in Fig.~\ref{fig6}. We observe that the iteration process is very important and our method converges within 6 iterations, and obtains the best performance on all datasets.
	

\subsection{Ablation Study}
We conduct an ablation study on miniImagenet, tieredImagenet, and CUB-200-2011, respectively, to assess the effects of the two components, \emph{i.e.}, learning to complete prototypes and Gaussian-based prototype fusion strategy. Specifically, in Table~\ref{table4}, (\romannumeral1) we remove all components, \emph{i.e.}, classifying each sample by the mean-based prototypes; (\romannumeral2) we add the ProtoComNet proposed in the conference version \cite{zhang2021prototype} (\emph{i.e.}, removing unseen parts/attributes) on (\romannumeral1) and classify each sample by the completed prototypes; (\romannumeral3) we extend \cite{zhang2021prototype} by introducing the PATNet on (\romannumeral2) to explore the unseen parts/attributes for ProtoComNet and classify each sample by the completed prototypes; (\romannumeral4) we fuse the mean-based and completed prototypes by MeanFusion; (\romannumeral5) we replace the MeanFusion of (\romannumeral4) by our two-step estimation method-based GaussFusion, i.e., the conference version \cite{zhang2021prototype}; (\romannumeral6) we replace the two-step estimation method by the improved EM-based estimation method on (\romannumeral5), where we don't use the EM-based estimation method because we have proved that the improved EM-based methods is more effective than EM-based methods in Tables~\ref{table1} and \ref{table2}.


\begin{table}[t]
	\caption{Ablation study on miniImagenet, tieredImagenet and CUB-200-2011. LCP: Learning to complete prototypes. GF, MF: Gaussian, mean-based prototype fusion. CV, EV: conference version \cite{zhang2021prototype}, extended version.}\smallskip
	\centering
	\smallskip\scalebox
	{0.9}{\begin{tabular}{c|c|c|c|c|c|c|c}
			\hline
			\multicolumn{1}{c|}{\multirow{2}{*}{}}& \multicolumn{2}{c|}{\multirow{1}{*}{LCP}} & \multicolumn{2}{c|}{\multirow{1}{*}{GF}} & \multicolumn{1}{c|}{\multirow{2}{*}{MF}} & \multicolumn{1}{c|}{\multirow{2}{*}{5-way 1-shot}} & \multicolumn{1}{c}{\multirow{2}{*}{5-way 5-shot}} \\
			\cline{2-5}
			& CV & EV & CV & EV & & \\
			\hline \hline
			\multicolumn{8}{c}{\multirow{1}{*}{miniImagenet}} \\
			\hline
			(\romannumeral1) & & & & &&61.22 $\pm$ 0.84$\%$ & 78.72 $\pm$ 0.60$\%$ \\
			(\romannumeral2) &$\surd$& & & && 65.62 $\pm$ 0.79$\%$ & 75.32 $\pm$ 0.61$\%$ \\
			(\romannumeral3) & &$\surd$ & & && 66.52 $\pm$ 0.84$\%$ & 75.68 $\pm$ 0.64$\%$ \\
			(\romannumeral4) & & $\surd$ & & & $\surd$ & 69.68 $\pm$ 0.76$\%$ & 81.65 $\pm$ 0.54$\%$\\
			(\romannumeral5) & & $\surd$ & $\surd$ & && 76.05 $\pm$ 0.79$\%$ & 83.32 $\pm$ 0.57$\%$ \\
			(\romannumeral6) & & $\surd$ & & $\surd$ && 79.01 $\pm$ 0.89$\%$ & 84.18 $\pm$ 0.56$\%$ \\
			\hline \hline
			\multicolumn{8}{c}{\multirow{1}{*}{tieredImagenet}} \\
			\hline
			(\romannumeral1)  & & & & &&69.02 $\pm$ 0.72$\%$ & $79.31 \pm 0.18\%$ \\
			(\romannumeral2) &$\surd$& & & && 71.66 $\pm$ 0.92$\%$ & $80.78 \pm 0.75\%$ \\
			(\romannumeral3) & &$\surd$ & & && 72.35 $\pm$ 0.90$\%$ & 84.10 $\pm$ 0.69$\%$ \\
			(\romannumeral4) & & $\surd$ & & & $\surd$ & 74.19 $\pm$ 0.90$\%$ & 86.09 $\pm$ 0.60$\%$ \\
			(\romannumeral5)& & $\surd$ & $\surd$ & &&  82.11 $\pm$ 0.90$\%$ & 88.46 $\pm$ 0.56$\%$ \\
			(\romannumeral6)& & $\surd$ & & $\surd$ && 83.06 $\pm$ 1.00$\%$ & 88.60 $\pm$ 0.57$\%$ \\
			\hline \hline
			\multicolumn{8}{c}{\multirow{1}{*}{CUB-200-2011}} \\
			\hline
			(\romannumeral1) & & & & &&77.75 $\pm$ 0.82$\%$ & $91.36 \pm 0.41\%$ \\
			(\romannumeral2) &$\surd$& & & && 84.36 $\pm$ 0.68$\%$ & $89.19 \pm 0.47\%$ \\
			(\romannumeral3) & &$\surd$ & & && 84.88 $\pm$ 0.68$\%$ & 89.51 $\pm$ 0.49$\%$ \\
			(\romannumeral4) & & $\surd$ & & & $\surd$ & 88.99 $\pm$ 0.58$\%$ & 94.05 $\pm$ 0.34$\%$\\
			(\romannumeral5)& & $\surd$ & $\surd$ & &&  93.30 $\pm$ 0.54$\%$ & 95.04 $\pm$ 0.30$\%$ \\
			(\romannumeral6)& & $\surd$ & & $\surd$ && 93.78 $\pm$ 0.55$\%$ & 95.19 $\pm$ 0.30$\%$ \\
			\hline
	\end{tabular}}
	\label{table4}
\end{table}

\emph{1)} {\bf Learning to Complete Prototypes.} From the results of (\romannumeral1) and (\romannumeral2) in Table~\ref{table4}, we observe that 1) the latter exceeds the former in 1-shot tasks, by around 4\%, which means that learning to complete prototypes is effective; 2) the latter obtains poor performance in 5-shot tasks. As our analysis in Section~\ref{section3_4}, the phenomenon results from the bias of ProtoComNet, namely the primitive knowledge noises or base-novel class differences. Besides, comparing the results of (\romannumeral2) and (\romannumeral3), we find that the latter achieves superior performance with an improvement of 1\% $\sim$ 2\%. This implies that exploiting unseen parts/attributes is effective and beneficial for estimating representative prototypes. 

\emph{2)} {\bf Gaussian-based Prototype Fusion Strategy.} According to the result in (\romannumeral4) and (\romannumeral5) of Table~\ref{table4}, we find that 1) the problem of ProtoComNet with poor performance on 5-shot tasks is effectively solved after we use the MeanFusion strategy (i.e., the assumption-based distribution estimation method); 2) the performance of the ProtoComNet can be further improved when it is combined with the GaussFusion with the two-step distribution estimation method, which is our conference strategy, by around 3\%. The result suggests that the two-step method is more effective than the assumption-based method. The key reason is the two-step method effectively estimates prototype distribution by exploiting the unlabelled samples. Besides, from the results of (\romannumeral5) and (\romannumeral6), we observe that the latter achieve 1\% $\sim$ 2\% higher classification accuracy. This is because the improved EM-based estimation method estimates more accurate prototype distribution for GaussFusion in an iterative manner.

Finally, to further verify that GaussFusion is able to alleviate the prototype completion error problem, we analyze the impacts of primitive knowledge with different noise levels $\gamma$ on classification performance. We report the results of miniImagenet, tieredImagenet, and CUB-200-2011 datasets in Table~\ref{table5}. Here, we introduce noises by randomly adding or removing class parts/attributes with probability $\gamma$. It can be observed that our method is more robust to primitive knowledge noises when GaussFusion is applied.

\begin{table}
	\caption{The performance analysis of primitive knowledge with different noise level $\gamma$ on 5-way 1-shot tasks of miniImagenet, tieredImagenet, and CUB-200-2011.}\smallskip
	\centering
	\smallskip\scalebox
	{1.0}{\begin{tabular}{l|c|c|c|c}
			\hline
			Methods & $\gamma=0.0$ & $\gamma=0.1$ & $\gamma=0.2$ & $\gamma=0.3$\\
			\hline \hline
			\multicolumn{5}{c}{\multirow{1}{*}{miniImagenet}} \\
			\hline
			w/o Fusion  & 65.99 $\%$ & 52.87 $\%$ & 46.68 $\%$ & 42.53 $\%$ \\
			w/ MeanFusion  & 69.64 $\%$ & 64.93 $\%$ & 60.28 $\%$ & 57.20 $\%$\\
			w/ GaussFusion  & 79.01 $\%$ & 77.89 $\%$ & 77.57 $\%$ & 77.24 $\%$\\
			\hline \hline
			\multicolumn{5}{c}{\multirow{1}{*}{tieredImagenet}} \\
			\hline
			w/o Fusion  & 72.35 $\%$ & 40.78 $\%$ & 32.37 $\%$ & 29.49 $\%$ \\
			w/ MeanFusion  & 74.19 $\%$ & 69.06 $\%$ & 62.93 $\%$ & 57.77 $\%$\\
			w/ GaussFusion  & 83.06 $\%$ & 81.60 $\%$ & 81.53 $\%$ & 81.51 $\%$\\
			\hline \hline
			\multicolumn{5}{c}{\multirow{1}{*}{CUB-200-2011}} \\
			\hline
			w/o Fusion  & 85.03 $\%$ & 82.24 $\%$ & 78.18 $\%$ & 74.17 $\%$ \\
			w/ MeanFusion  & 85.34 $\%$ & 85.02 $\%$ &84.66 $\%$ & 84.18 $\%$\\
			w/ GaussFusion  & 93.78 $\%$ & 93.65 $\%$ & 93.56 $\%$ & 93.28 $\%$\\
			\hline
	\end{tabular}}
	\label{table5}
\end{table} 

\subsection{Visualization}
In this subsection, we conduct visualization analysis on feature space  to answer the following two questions:

{\bf How are the part/attribute distributed in the feature space?} To understand how our method complete prototypes by using extracted part/attribute features, we randomly select two part/attribute from miniImagenet, \emph{i.e.}, ``paw'' and ``tail''. We visualize of all classes by t-SNE in the feature space, where the classes with the part/attribute ``paw'' or ``tail'' are marked in color ``red'', otherwise in color ``blue''. As shown in Fig.~\ref{fig9}, we find these classes that have the same attributes are clustered together, which is beneficial to learn to complete prototypes.

{\bf How does our method work?} To understand how does the proposed method work, we select a 5-way 1-shot and 5-shot classification task from the meta-test set of miniImageNet to visualize the prototypes and samples by t-SNE. As shown in Fig.~\ref{fig10}, after completing and fusing the class prototypes, the fused prototypes (marked in squares) become closer to real prototypes (marked in stars).

\begin{figure}[!t]
	\centering
	\subfigure[miniImagenet (part: ``paw'')]{ 
		\label{fig9a} 
		\includegraphics[width=0.48\columnwidth]{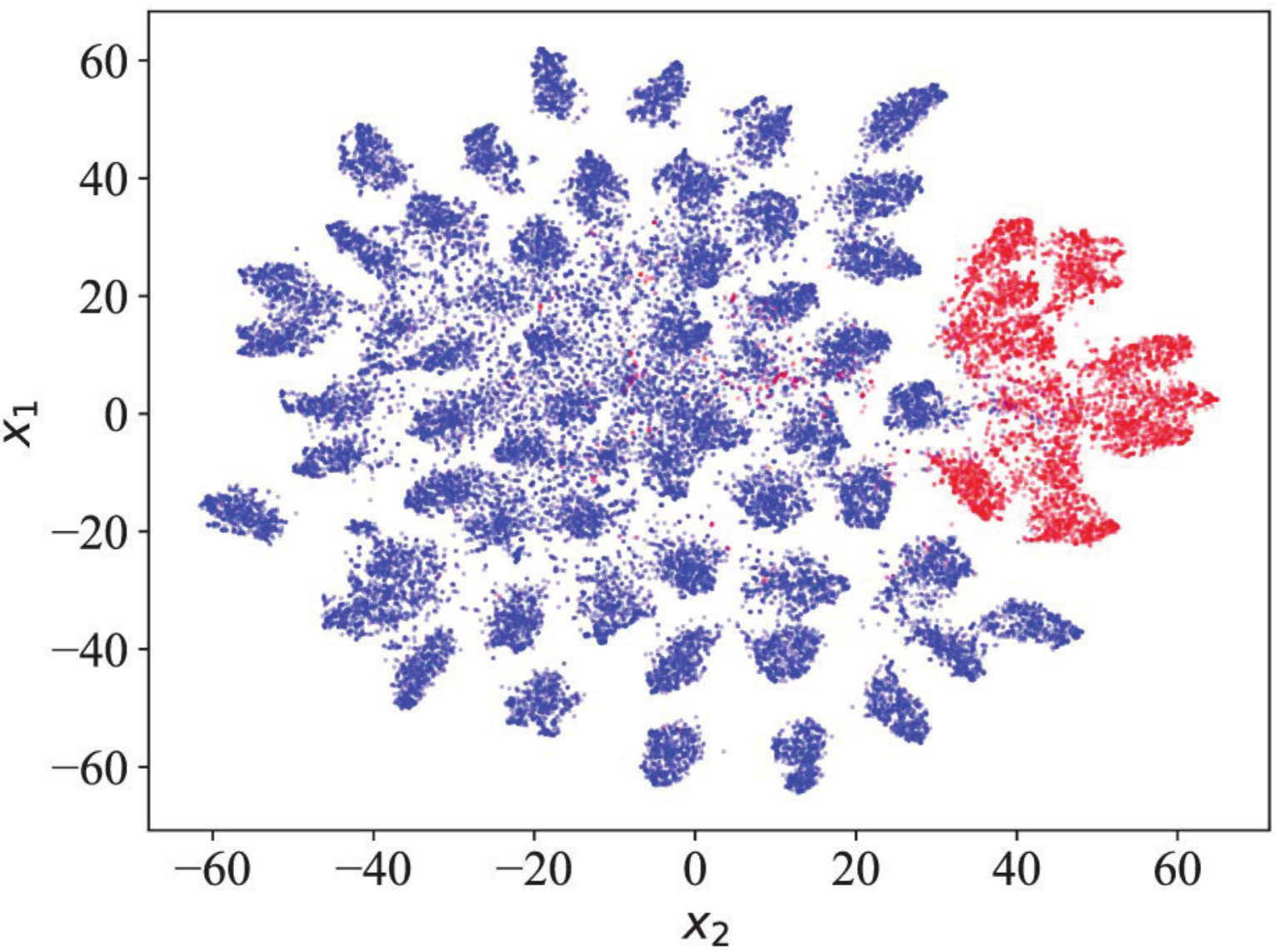}}
	\subfigure[miniImagenet (part: ``tail'')]{ 
		\label{fig9b} 
		\includegraphics[width=0.48\columnwidth]{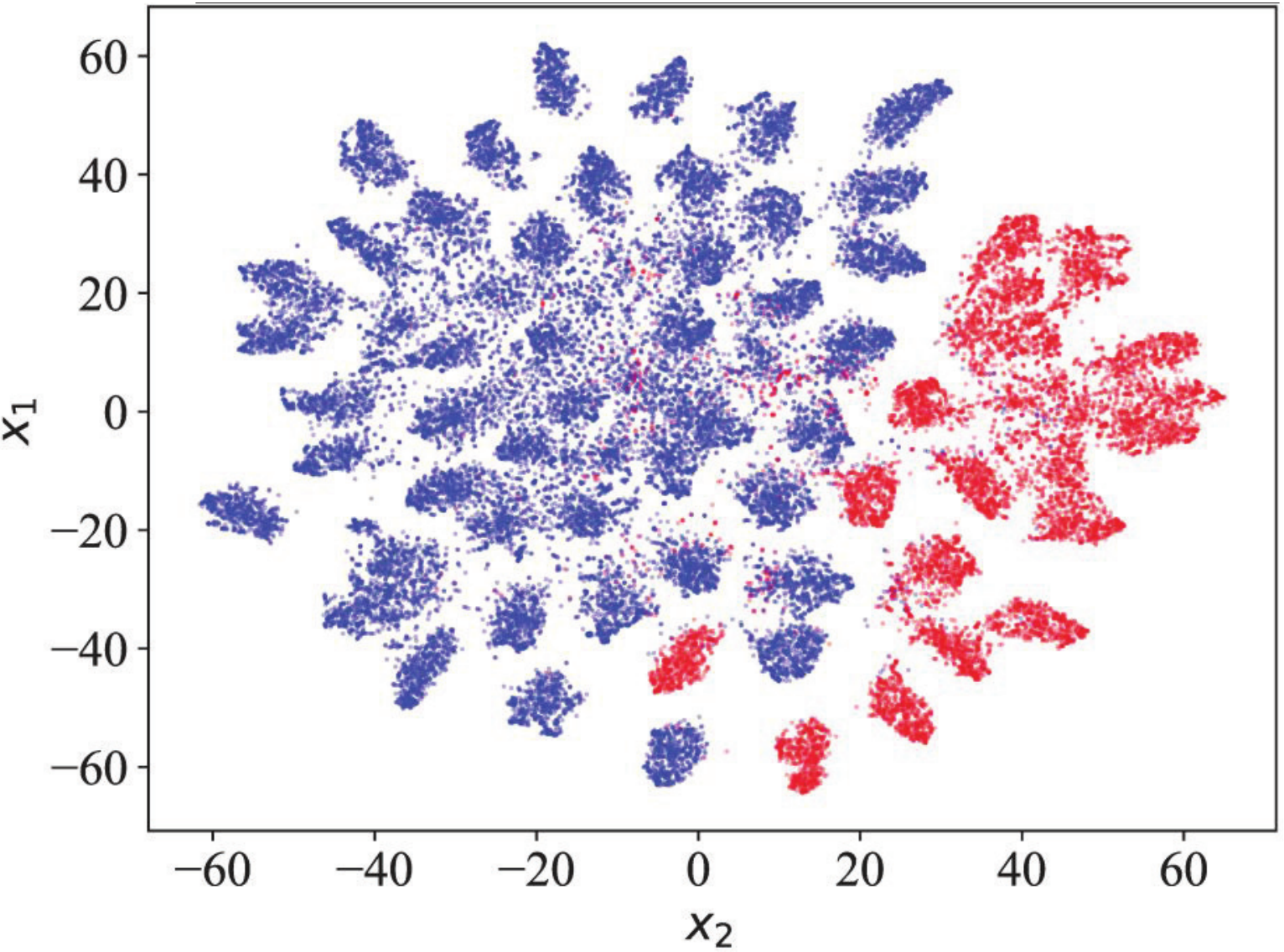}}
	\caption{Visualization of part/attribute feature on miniImageNet.}
	\label{fig9}
\end{figure} 

\begin{figure}[!t]
	\centering
	\subfigure[5-way 1-shot task]{ 
		\label{fig10a} 
		\includegraphics[width=0.48\columnwidth]{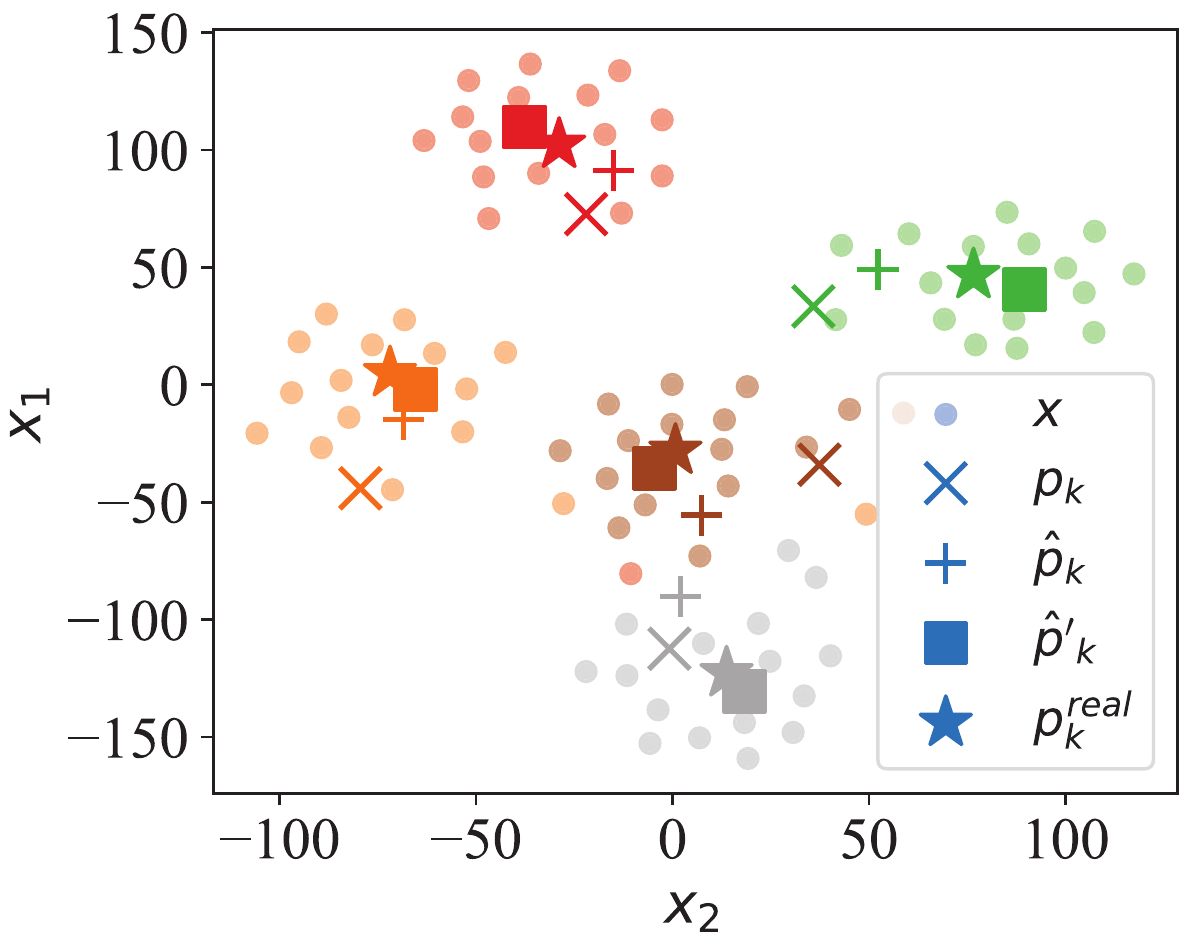}}
	\subfigure[5-way 5-shot task]{ 
		\label{fig10b} 
		\includegraphics[width=0.48\columnwidth]{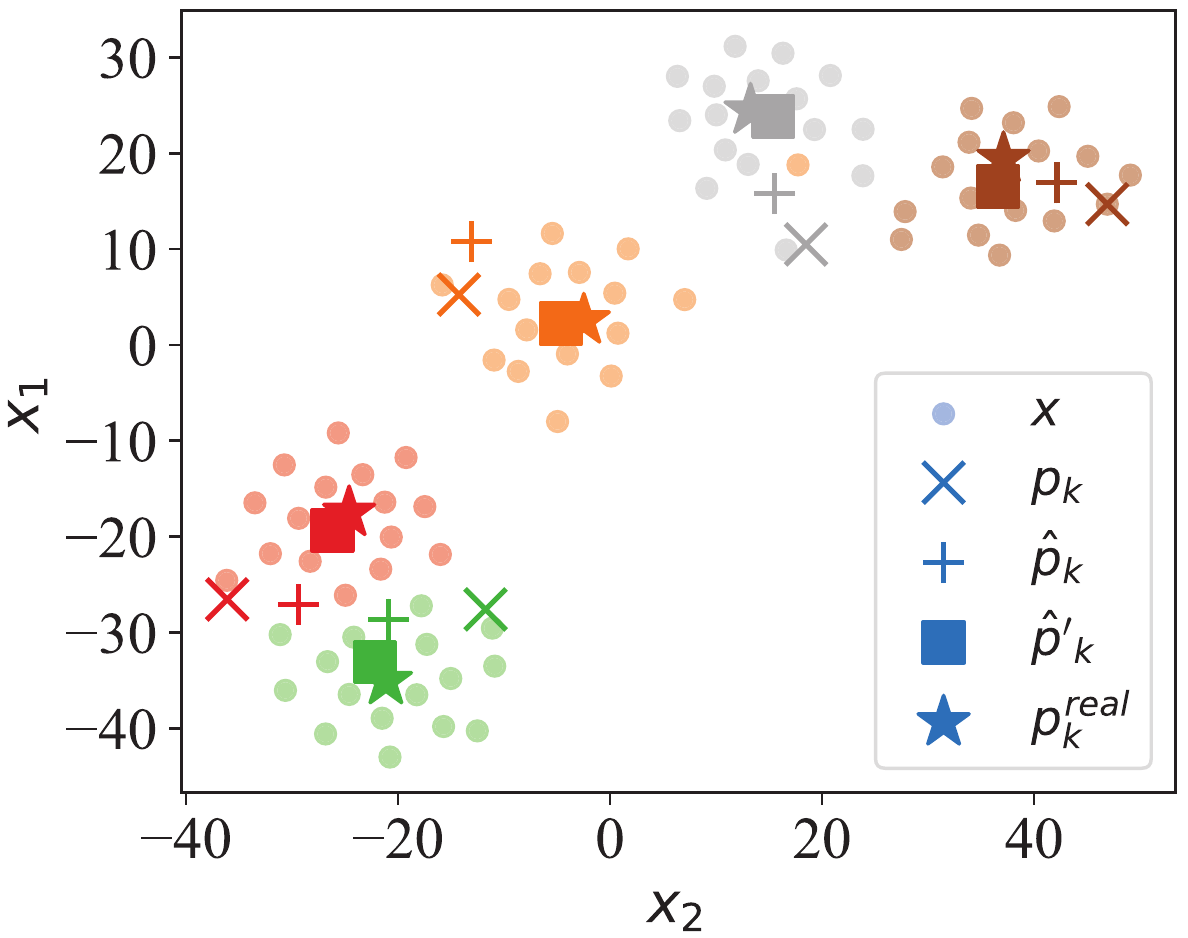}}
	\caption{Visualization of a 5-way 1/5-shot task sampled from the meta-test set of miniImageNet. Best viewed in color.}
	\label{fig10}
\end{figure} 

\section{Conclusion}
\label{section_5}
For few-shot learning, a simple pre-training on base classes can obtain a good feature extractor, where the novel class samples can be well clustered together. The key challenge is how to obtain more representative prototypes because the novel class samples spread as groups with large variances. To solve the issue, we introduce primitive knowledge and extract representative feature for seen attribues as priors. Then we propose a part/attribute transfer network to infer the visual features for unseen parts/attributes as supplementary priors, a prototype completion network to complete prototypes via primitive knowledge and these priors, and a Gaussian-based prototype fusion strategy to alleviate the prototype completion error problem. Particularly, in the fusion strategy, we develop three methods to estimate fusion parameters, i.e., two-step method, EM (Expectation Maximization)-based method, and improve EM-based estimation method. Experiments show that our method obtains superior performance on three benchmark data sets.


%

\ifCLASSOPTIONcompsoc
  \section*{Acknowledgments}
\else
  \section*{Acknowledgment}
\fi

This work was supported by the Shenzhen Science and Technology Program under Grant No. JCYJ201805071838- 23045 and Grant No. JCYJ20200109113014456.

\ifCLASSOPTIONcaptionsoff
  \newpage
\fi



\bibliographystyle{IEEEtran}
\bibliography{egbib}

\begin{IEEEbiographynophoto}{Baoquan Zhang}
	is currently pursuing the Ph.D. degree with the School of Computer Science and Technology, Harbin Institute of Technology, Shenzhen, China. His current research interests include meta learning, few-shot learning, and machine learning.
\end{IEEEbiographynophoto}
\begin{IEEEbiographynophoto}{Xutao Li}
	is currently an Associate Professor with the School of Computer Science and Technology, Harbin Institute of Technology, Shenzhen, China. His research interests include data mining, machine learning, graph mining, and social network analysis, especially tensor-based learning, and mining algorithms.
\end{IEEEbiographynophoto}
\begin{IEEEbiographynophoto}{Yunming Ye}
	is currently a Professor with the School of Computer Science and Technology, Harbin Institute of Technology, Shenzhen, China. His research interests include data mining, text mining, and ensemble learning algorithms.
\end{IEEEbiographynophoto}
\begin{IEEEbiographynophoto}{Shanshan Feng}
	is currently an Associate Professor with the School of Computer Science and Technology, Harbin Institute of Technology, Shenzhen, China. His research interests include sequential data mining and social network analysis.
\end{IEEEbiographynophoto}

\appendices
\section{Derivation of GaussFusion}
\label{appendix_a}
\noindent {\bf Proposition.} Let $f(x)$ and $g(x)$ be a Multivariate Gaussian Distributions with diagonal covariance, \emph{i.e.}, $f(x) = N(\hat{\mu}_{k}, diag(\hat{\sigma}_{k}^2))$ and $g(x) = N(\mu_{k}, diag(\sigma_{k}^2))$ where $x$ is a $d$-dimension random vector, $\hat{\mu}_{k}$ and $\mu_{k}$ denote $d$-dimension mean vector, and $\hat{\sigma}_{k}^2$ and $\sigma_{k}^2$ are $d$-dimension variance vector. Then, their product obeys a new Multivariate Gaussian Distributions $N(\mu'_{k}, diag({\sigma'_{k}}^2))$ with $\mu'_{k}=\frac{\sigma_{k}^2 \odot \hat{\mu}_{k} + \hat{\sigma}_{k}^2 \odot \mu_{k}}{\hat{\sigma}_{k}^2 + \sigma_{k}^2}$ and ${\sigma'_{k}}^2=\frac{\sigma_{k}^2 \odot \hat{\sigma}_{k}^2}{\hat{\sigma}_{k}^2 + \sigma_{k}^2}$, where $\odot$ denotes the element-wise product.

\noindent {\bf Derivation.} Considering that the covariances of $f(x)$ and $g(x)$ are simplified as diagonal covariances. This means that the variables of the random vector $x$ are uncorrelated. In this case, $f(x)$ and $g(x)$ can be simplified as the expression below:

\begin{equation} 
\nonumber 
f(x)=\prod_{i=0}^{d-1} \frac{1}{\sqrt{2\pi{\hat{\sigma}_{k,i}}^2}}\ e^{(\frac{-(x_i-\hat{\mu}_{k,i})^2}{2\hat{\sigma}_{k,i}^2})}
\end{equation}
\begin{equation} 
\nonumber 
g(x)=\prod_{i=0}^{d-1} \frac{1}{\sqrt{2\pi{\sigma^2_{k,i}}}}\ e^{(\frac{-(x_i-\mu_{k,i})^2}{2\sigma_{k,i}^2})}
\end{equation}
Thus, their product $h(x)$ satisfies:
\begin{equation}
\nonumber 
\begin{aligned}
&h(x) \\&= f(x)g(x) 
\\& = \prod_{i=0}^{d-1} \frac{1}{\sqrt{2\pi\hat{\sigma}^2_{k,i}}}\ e^{(\frac{-(x_i-\hat{\mu}_{k,i})^2}{2\hat{\sigma}_{k,i}^2})} \ \frac{1}{\sqrt{2\pi\sigma^2_{k,i}}}\ e^{(\frac{-(x_i-\mu_{k,i})^2}{2\sigma_{k,i}^2})}
\\& = \prod_{i=0}^{d-1} \frac{1}{2\pi\sqrt{\hat{\sigma}^2_{k,i}\sigma^2_{k,i}}}\ e^{(\frac{-(x_i-\hat{\mu}_{k,i})^2}{2\hat{\sigma}_{k,i}^2} + \frac{-(x_i-\mu_{k,i})^2}{2\sigma_{k,i}^2})}
\\& = \prod_{i=0}^{d-1} \frac{1}{2\pi\sqrt{\hat{\sigma}^2_{k,i}\sigma^2_{k,i}}}\ e^{(\frac{(x_i-\frac{\sigma_{k,i}^2\hat{\mu}_{k,i} + \hat{\sigma}_{k,i}^2\mu_{k,i}}{\hat{\sigma}_{k,i}^2 + \sigma_{k,i}^2})^2}{2\frac{\sigma_{k,i}^2 \hat{\sigma}_{k,i}^2}{\hat{\sigma}_{k,i}^2 + \sigma_{k,i}^2}} + \frac{(\hat{\mu}_{k,i} - \mu_{k,i})^2}{2(\hat{\sigma}_{k,i}^2 + \sigma_{k,i}^2)})}
\\&= \prod_{i=0}^{d-1} \frac{S_{i}}{\sqrt{2\pi\frac{\sigma_{k,i}^2 \hat{\sigma}_{k,i}^2}{\hat{\sigma}_{k,i}^2 + \sigma_{k,i}^2}}} e^{(-\frac{-(x_i-\frac{\sigma_{k,i}^2\hat{\mu}_{k,i} + \hat{\sigma}_{k,i}^2\mu_{k,i}}{\hat{\sigma}_{k,i}^2 + \sigma_{k,i}^2})^2}{2(\frac{\sigma_{k,i}^2 \hat{\sigma}_{k,i}^2}{\hat{\sigma}_{k,i}^2 + \sigma_{k,i}^2})})}
\end{aligned}
\end{equation}
where $S_{i}=\frac{1}{\sqrt{2\pi(\sigma_{k,i}^2 + \hat{\sigma}_{k,i}^2)}} e^{-\frac{(\hat{\mu}_{k} - \mu_{k})^2}{2(\hat{\sigma}_{k}^2 + \sigma_{k}^2)}}$. Thus, $h(x)$ is also a multivariate Gaussian distribution, \emph{i.e.}, $N(\mu'_{k}, diag({\sigma'_{k}}^2))$ with mean $\mu'_{k}=\frac{\sigma_{k}^2 \odot \hat{\mu}_{k} + \hat{\sigma}_{k}^2 \odot \mu_{k}}{\hat{\sigma}_{k}^2 + \sigma_{k}^2}$ and diagonal covariance $diag({\sigma'_{k}}^2)$ where ${\sigma'_{k}}^2=\frac{\sigma_{k}^2 \odot \hat{\sigma}_{k}^2}{\hat{\sigma}_{k}^2 + \sigma_{k}^2}$.

\section{Workflow of Improved EM-based Estimation Method} 
\label{appendix_b}
In the section, we provide implementation details of the improved EM-based estimation method for reproducibility. The overall workflow is summarized in Algorithm~\ref{algorithm1}. Specifically, given the support set $\mathcal{S}$,  the query set $\mathcal{Q}$, the mean-based prototypes $p_k$, and the completed prototypes ${\hat p}_k$, we perform the following four steps to estimate the prototype fusion parameters for GaussFusion: (1) initilizing the mean $\mu_{k}$ or $\hat{\mu}_{k}$ by using the prototypes $p_k$ or ${\hat p}_k$ (Line 1); (2) performing the E-Step to compute the posterior probability that a given sample $x \in \mathcal{Q}$ belongs to a given class $k$ by following Eq.~(\ref{eq12}) (Line 3). (3) performing the M-Step to obtain the optimal mean and diagonal covariance $\mu_{k}$ and $\sigma_{k}$ or $\hat{\mu}_{k}$ and $\hat{\sigma}_{k}$ by Eqs.~(\ref{eq13}) and (\ref{eq14}) (Line 4); (4) Repeatly performing the step (1) and (2) until the maximum number of iterations $n_{iter}$ is reached (Lines 2 - 6).

\renewcommand{\algorithmicrequire}{\textbf{Input:}}
\renewcommand{\algorithmicensure}{\textbf{Inference:}}
\begin{algorithm}[htbp]
	\caption{Improved EM-based estimation method}
	\label{algorithm1}
	\begin{algorithmic}[1]
		\REQUIRE ~~\\ 
		A support set  $\mathcal{S}=\{(x_i, y_i)\}_{i=0}^{N \times K}$, a query set $\mathcal{Q}=\{(x_i, y_i)\}_{i=0}^{M}$ from novel classes, the feature extractor $f_{\theta_f}()$, and the prototypes $p_k$ or ${\hat p}_k$.
		\renewcommand{\algorithmicrequire}{\textbf{Output:}}
		\REQUIRE ~~\\ 
		The estimated mean and diagonal covariance of prototype distribution, \emph{i.e.}, $\mu_{k}$ and $\sigma_{k}$, or $\mu'_{k}$ and $\sigma'_{k}$.
		\renewcommand{\algorithmicrequire}{\textbf{Initialization:}}
		\ENSURE ~~\\
		\STATE Initilizing $\mu_{k}$ or $\hat{\mu}_{k}$ with $p_k$ or ${\hat p}_k$ and regarding them as initial prototypes of cosine classifier;
		\FOR {$t = 0, 1, ..., n_{iter}-1$}
		\STATE {\bf E-Step.} Estimating the posterior probability that a given sample $x \in \mathcal{S}/\mathcal{Q}$ belongs to a given class $k$ with Eq.~(\ref{eq12}) or one-hot vector of its labels;  
		\STATE {\bf M-Step.} Estimating the mean and diagonal covariance $\mu_{k}$ and $\sigma_{k}$ or $\hat{\mu}_{k}$ and $\hat{\sigma}_{k}$ with Eqs.~(\ref{eq13}) and (\ref{eq14}) to maximize the posterior probability; 
		\STATE Replacing the above prototypes by mean $\mu_{k}$ or $\mu'_{k}$; 
		\ENDFOR 
	\end{algorithmic}
\end{algorithm}

\section{Detailed Information of Three Datasets} 
\label{appendix_c}
We summarize the necessary information about the three data sets in Table~\ref{table0}. Note that different from our conference version \cite{zhang2021prototype}, the extended method can exploit unseen parts/attributes of novel classes for prototype completion.

\begin{table}[htbp]
	\caption{The basic statistics of the MiniImagenet, TieredImagenet and CUB-200-2011 dataset.}\smallskip
	\centering
	\smallskip\scalebox
	{0.95}{\begin{tabular}{l|c|c|c|c|c|c}
			\hline
			\multicolumn{1}{c|}{\multirow{2}{*}{Datasets}} & \multicolumn{3}{c|}{\multirow{1}{*}{Number of Class}} & \multicolumn{3}{c}{\multirow{1}{*}{Number of Part/attribute}}\\
			\cline{2-7}
			& train & val & test & seen & unseen & all\\
			\hline
			\hline
			MiniImagenet & 64 & 16 & 20 & 168 & 122 & 290\\
			TieredImagenet & 351 & 97 & 160 & 411 & 165 & 576\\
			CUB-200-2011 & 100 & 50 & 50 & 171 & 141 & 312\\
			\hline
	\end{tabular}}
	\label{table0}
\end{table}

\section{Additional Visualization} 
\label{appendix_d}
\noindent {\bf Is reasonable our motivation on other data sets?} To further verify the reasonability of our motivation (i.e., estimating more accurate prototypes is more effective than fine-tuning feature extractor during meta-learning), we additionally visualize the distribution of base and novel class samples of the tieredImagenet and CUB-200-2011 data sets in the pre-trained feature space in Fig.~\ref{fig11} and \ref{fig12}. Note that we randomly select 15\% of the classes from the base and novel classes on tieredImagenet for clarity. We have the similar observations as those in Fig.~\ref{fig1} of Section~\ref{sec:introduction}, that is, the base class samples form compact clusters while the novel class samples spread as groups with large variances. This means that our motivation is reasonable and the problem of inaccurate estimation of prototypes widely exists in the pre-trained feature space for the real-world data sets.

\begin{figure}[h]
	\centering
	\subfigure[Base Classes ($\mathrm{\sigma^2=0.36}$)]{ 
		\label{fig11a} 
		\includegraphics[width=0.45\columnwidth]{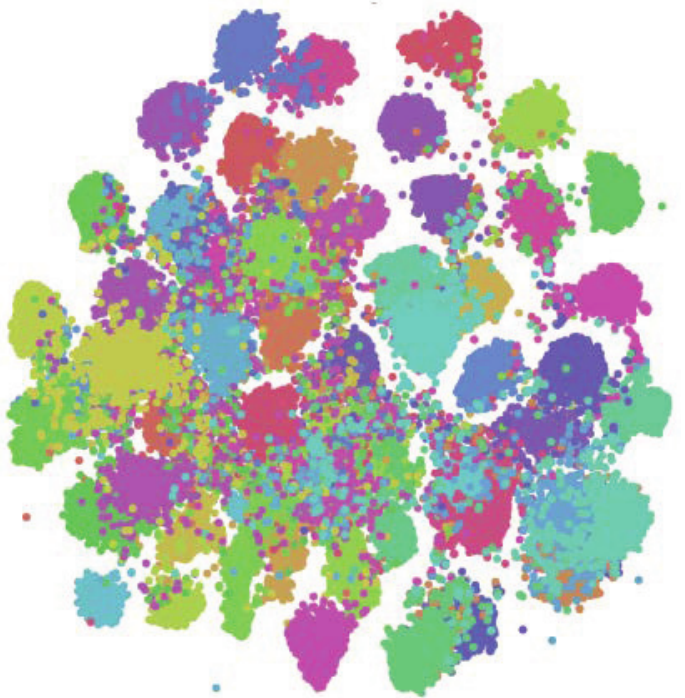}}
	\subfigure[Novel Classes ($\mathrm{\sigma^2=0.41}$)]{ 
		\label{fig11b} 
		\includegraphics[width=0.48\columnwidth]{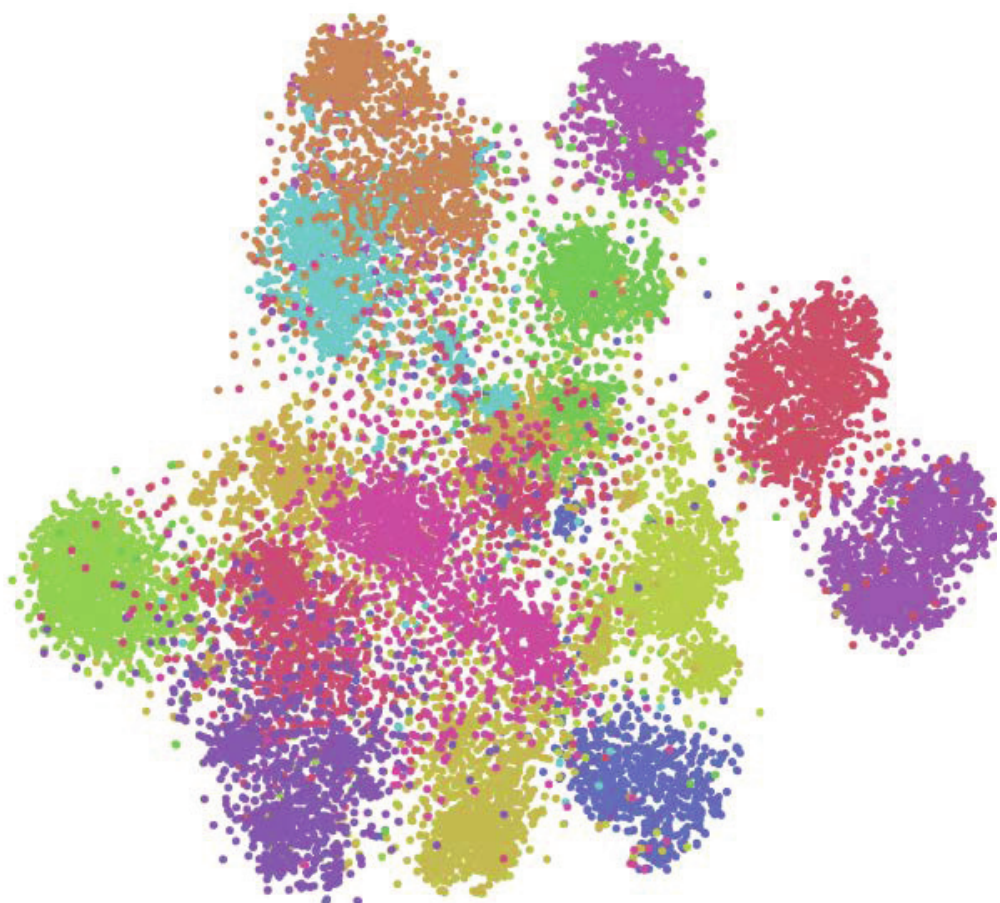}}
	\caption{The distribution of base and novel class samples in the pre-trained feature space on tieredImagenet data set.  ``$\mathrm{\sigma}^2$'' denotes the averaged variance. }
	\label{fig11}
\end{figure}

\begin{figure}[h]
	\centering
	\subfigure[Base Classes ($\mathrm{\sigma^2=0.054}$)]{ 
		\label{fig12a} 
		\includegraphics[width=0.45\columnwidth]{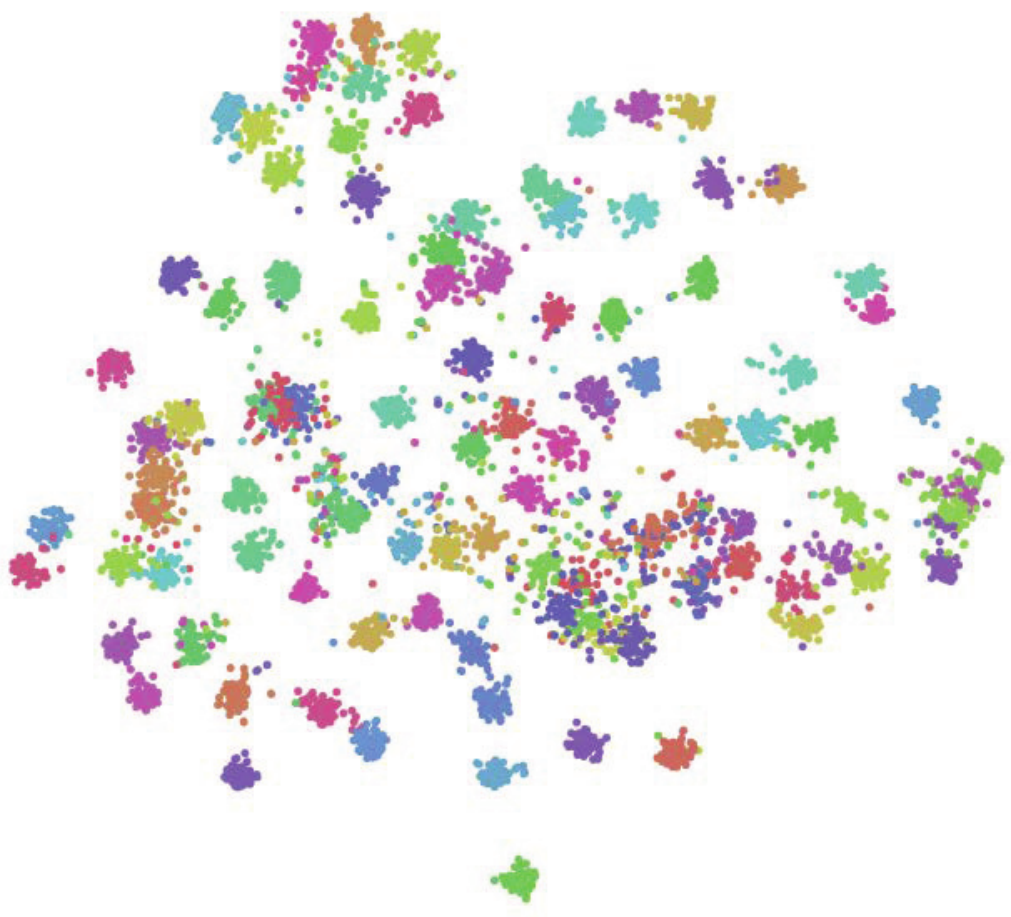}}
	\subfigure[Novel Classes ($\mathrm{\sigma^2=0.063}$)]{ 
		\label{fig12b} 
		\includegraphics[width=0.48\columnwidth]{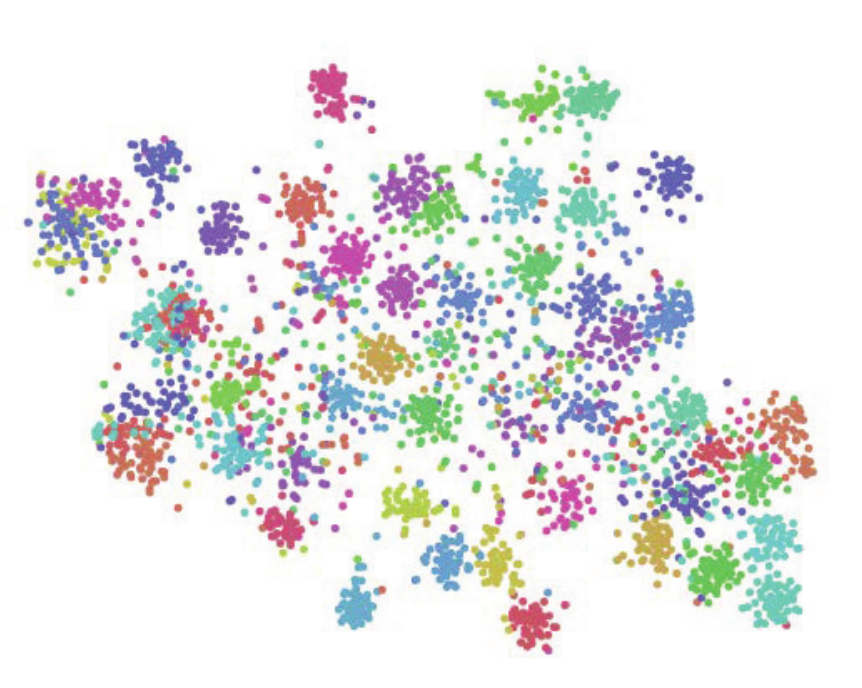}}
	\caption{The distribution of base and novel class samples in the pre-trained feature space on CUB-200-2011 data set.  ``$\mathrm{\sigma}^2$'' denotes the averaged variance. }
	\label{fig12}
\end{figure}

\end{document}